\documentclass[runningheads]{llncs} 
\usepackage{graphicx}
\usepackage{amsmath}
\DeclareMathOperator*{\argmax}{arg\!\max}

\usepackage[table,xcdraw]{xcolor}

\usepackage{tikz}
\usepackage{comment}
\usepackage{amsmath,amssymb} %
\usepackage{color}

\usepackage{xspace}
\newcommand*{\eg}{e.g.\@\xspace}
\newcommand*{\ie}{i.e.\@\xspace}

\usepackage{todonotes}
\usepackage{subfig}
\usepackage{hyperref}
\hypersetup{
    colorlinks=true,
    linkcolor=black,
    filecolor=magenta,      
    urlcolor=cyan,
    pdftitle={Overleaf Example},
    pdfpagemode=FullScreen,
    }
\usepackage{xr-hyper}

\begin{document}
\pagestyle{headings}
\mainmatter
\def\ECCVSubNumber{6560}  %

\title{Visual Knowledge Tracing}

\titlerunning{Visual Knowledge Tracing}

\author{
{Neehar Kondapaneni\inst{1} \quad Pietro Perona\inst{1} \quad Oisin Mac Aodha\index{Mac Aodha, Oisin}\inst{2}} \\ 
}

\authorrunning{Kondapaneni et al.}
\institute{$^1$Caltech \quad $^2$University of Edinburgh}

\maketitle
\begin{abstract}
Each year, thousands of people learn new visual categorization tasks -- radiologists learn to recognize tumors, birdwatchers learn to distinguish similar species, and crowd workers learn how to annotate valuable data for applications like autonomous driving. 
As humans learn, their brain updates the visual features it extracts and attend to, which ultimately informs their final classification decisions. 
In this work, we propose a novel task of tracing the evolving classification behavior of human learners as they engage in challenging visual classification tasks. 
We propose models that jointly extract the visual features used by learners as well as predicting the classification functions they utilize. 
We collect three challenging new datasets from real human learners in order to evaluate the performance of different visual knowledge tracing methods. 
Our results show that our recurrent models are able to predict the classification behavior of human learners on three challenging medical image and species identification tasks. 

\end{abstract}

\section{Introduction}  

Humans excel at learning new concepts even when they have only received limited explicit supervision~\cite{markman1989categorization,xu2007word}. 
Key to our success is our ability to extract informative and generalizable representations from the world around us and our ability to update these representations given relatively sparse feedback. 
This capacity, in turn, enables us to perform complex tasks such as spatial navigation and visual categorization with apparent ease.    

Despite recent progress that has been made in computer vision in learning visual representations through self-supervision alone~\cite{wu2018unsupervised,chen2020simple,he2020momentum}, large amounts of supervision are still required to make best use of the resulting features~\cite{cole2021does}. 
In light of this, it is important for us to better understand: (i) what are the  properties that make representations learned by \emph{humans} so effective,  (ii) how are these representations learned, and (iii) can we predict the classification behavior of humans during learning?  
Our ultimate goal is to obtain better insight into how humans are such effective learners, which can then potentially inform new learning mechanisms for future artificial systems.  

\begin{figure}[t]
 \begin{minipage}[c][50pt][c]{0.48\textwidth}
    \includegraphics[trim={110pt 30pt 110pt 30pt},clip,width=\linewidth]{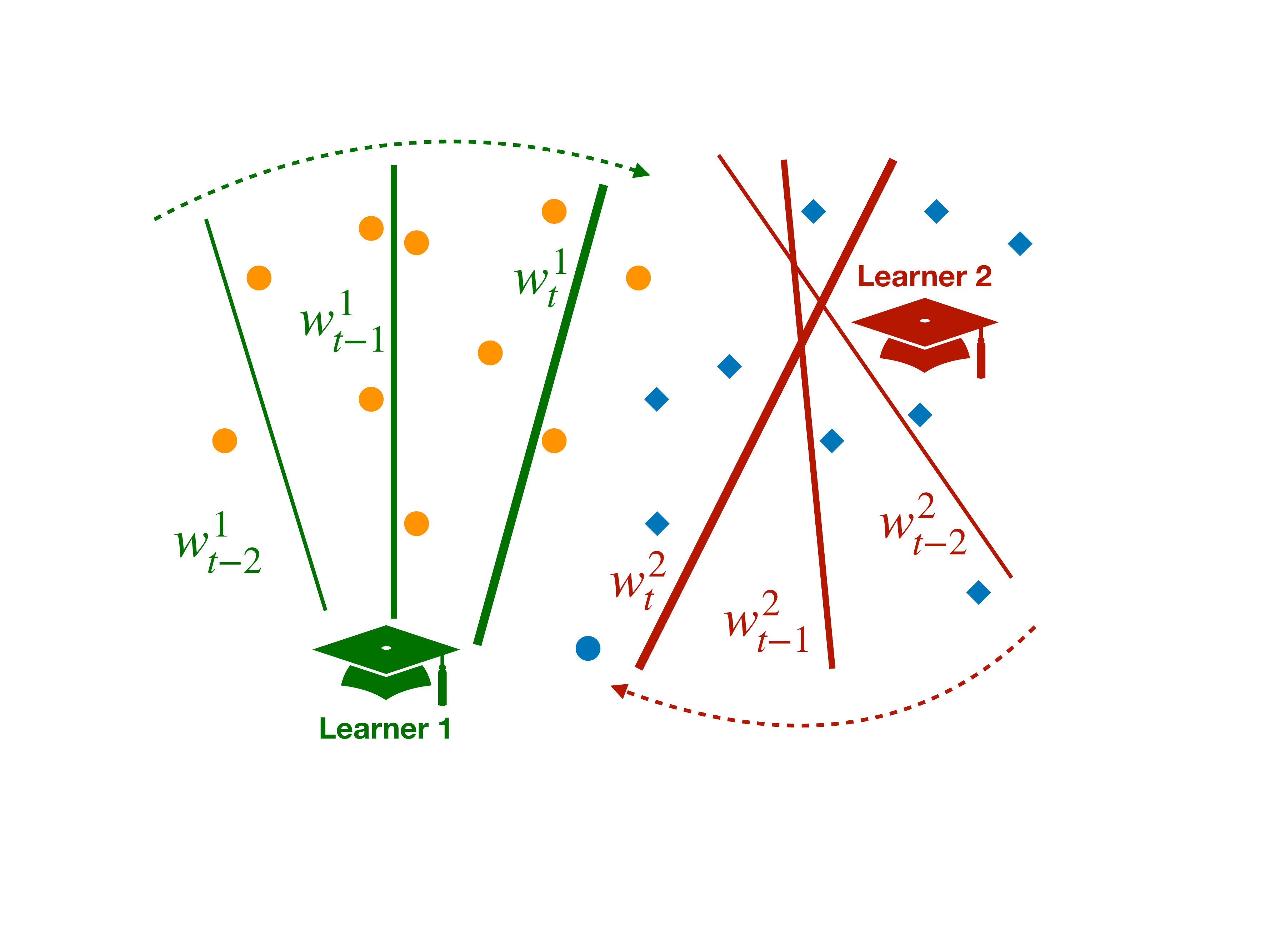}
  \end{minipage} %
  \begin{minipage}[l][40pt][c]{0.5\textwidth}
    \caption{\footnotesize
      We model a learner as an evolving classifier in a learned feature space. We assume the feature space is static and, as the learner is presented with images and class labels over time, their internal classification function self-updates. 
      Here, we illustrate this for three time-steps, for two learners, learning a binary classification task (orange versus blue). 
    } \label{fig:overview_fig}
  \end{minipage}
\end{figure}

In order to attempt to address some of these questions, in this work we explore the problem of visual knowledge tracing. 
In the educational data mining community, knowledge tracing is the problem of monitoring and predicting the evolving knowledge state of a learner engaged in a learning task~\cite{liu2021survey}. 
Recent work has applied advances in deep learning to knowledge tracing for question and answer-style text datasets and has investigated applications in domains such as mathematics education~\cite{piech2015deep,pandey2019self,pu2020deep}. 
However, tracing the behavior of humans that are engaged in learning and performing challenging visual categorization tasks is underexplored. 
Most closely related is the work on deep metric learning that attempts to learn human aligned visual representations from sparse human annotations~\cite{kaya2019deep}. 
However, these works typically make simplifying assumptions, \eg assuming the learners are not changing over time (\ie they are `static') or that the visual criteria used by the learners is the same across all learners. %
Instead, we explore a more challenging visual knowledge tracing setting where the learners are assumed to be non-stationary during learning, \ie the visual features they use to perform the classification task at hand can, and likely do, change over time (see Fig.~\ref{fig:overview_fig}).

We present a recurrent neural network-based approach for visual knowledge tracing.  
Once trained, our models are capable of predicting the classification behavior of human learners that were not observed during training. 
The proposed models make use of the history of previous learner responses, images, and ground truth class labels in order to predict their future responses. 
Through experiments on three challenging image classification datasets we show that our models are superior to baseline approaches. %
Our models are capable of tracing the learning dynamics more accurately than non-recurrent baselines.

We make the following three contributions: 
(i) A new model for visual knowledge tracing that jointly estimates the visual features and per-time-step classification function used by non-stationary human learners.   
(ii) A new set of annotations for three benchmark evaluation datasets collected from humans engaged in learning challenging visual classification tasks.    
(iii) A detailed comparison of several visual knowledge tracing methods on these datasets.

\section{Related Work}

\subsection{Metric Learning} 
The goal of metric learning is to learn perceptual embeddings such that distance in the lower dimensional embedding space encodes information related to semantic similarity. 
Pre-deep learning approaches to metric learning were primarily concerned with learning embeddings directly for each item in an input set. 
In the case of learning from human supervision, approaches that use relative similarity judgements have been shown to be effective~\cite{kruskal1978multidimensional,van2012stochastic,wilber2014cost,roads2021predicting}. 
These methods have also been extended to the adaptive setting where the model can decide which items to request annotations for in order to speed up training~\cite{tamuz2011adaptively}. 

More recently, end-to-end metric learning methods have attempted to parameterize an embedding function (\eg a convolutional neural network) directly~\cite{roth2020revisiting,musgrave2020metric}. 
As a result, they are able to embed any new item into the embedding space, even those not seen at training time. 
Representative earlier applications of this line of work include image ranking~\cite{wang2014learning} and  face recognition~\cite{schroff2015facenet}. 
The standard assumption made by the majority of these methods is that only \emph{one} similarity criterion is being used. 
However, when collecting data from human annotators, different individuals may be using different visual criteria when making classification and similarity judgements, \eg one individual could be using shape, while the other is using color. 
Furthermore, the same individual may change the criteria they use conditioned on the specific items they are shown, and could change to another criteria when shown another set of items at a different point in time.

There has been some work that attempts to deal with the fact that \emph{different} similarity criteria may be being used. 
These range from fully supervised, where the similarity criteria is known at training time~\cite{veit2017conditional}, through to unsupervised methods that attempt to estimate the criteria~\cite{kim2018context,nigam2019towards,tan2019learning,welinder2010multidimensional}.
Attempts have also been made to probabilistically estimate item  embeddings along with annotator-specific parameters representing the individual criteria they are using~\cite{welinder2010multidimensional}. 

One common assumption made by the above methods is that the learner is stationary, \ie they are used a predefined and fixed similarity criteria, or small set of criteria, which do not evolve or change over time. 
This is a reasonable assumption to make when dealing with common everyday object categories where the annotators will likely be familiar with the objects depicted and have an a priori understanding of how the visual features of the objects may vary. 
However, this assumption is violated in cases where the annotator is in the process of \emph{learning} the visual concepts of interest. 
In this work, we address this non-stationary setting and show that by doing this, we can more accurately predict the visual classification behavior of real human learners.

\subsection{Human Category Representation} 
Existing models of human category representation and learning can be clustered into five major groups: rule-based, prototype-based, exemplar-based, knowledge / theory-based, and decision boundary-based approaches~\cite{zeithamova2008category,ashby2011human}. 
The current consensus is that humans likely use multiple different category-learning systems depending on the specific nature of the task at hand~\cite{ashby2005humanmult,ashby2005human}. 
For example, in rule-based tasks, the optimal policy may be easy to verbalize and thus efficiently encoded via a set of rules. 
In practice however, perceptual tasks such as fine-grained visual categorization can be much harder to represent in this way~\cite{biederman1987sexing}.
Several works have attempted to extract perceptual embeddings that align with human similarity judgements from coarse~\cite{hebart2020revealing,roads2021enriching} and more fine-grained~\cite{nosofsky2018toward} image collections. 
\cite{attarian2020transforming} showed that with simple linear transformations, pre-trained deep image classifiers can be predictive of human similarity judgements. 
Relevant to our work, \cite{barry2021human} investigated whether human learning dynamics mimic gradient descent in Artificial Neural Networks when learning visual categories. 
However, they assumed they had access to the feature space used by the human, whereas we instead attempt to learn this. 
In this work, we also aim to extract human-aligned representations, but in the more challenging setting whereby our learners are not static, but instead are in the process of learning the categorization task. 

\subsection{Knowledge Tracing} 
The problem of modelling the hidden state of dynamic learners as they interact with a learning task has also been tackled in the knowledge tracing literature. 
Bayesian Knowledge Tracing-based methods model a learner's knowledge state by assuming that the learner can be represented as a Markov process which updates online during learning~\cite{corbett1994knowledge}.  
Building on this line of work, Deep Knowledge Tracing (DKT) instead uses a recurrent neural network as the underlying tracing model~\cite{piech2015deep}, and fully self-attention-based methods have also been proposed~\cite{pandey2019self,pu2020deep}. 
It is important to note that conventional knowledge tracing attempts to model knowledge acquisition as a binary variable at the `skill' level (\ie the visual class) as opposed to the `instance' level (\ie a specific image). 
In contrast, our approach jointly learns an image embedding function in addition to being able to capture and predict the individual learning trajectories of multiple different learners. A detailed comparison of the original DKT model and our model is provided in the supplementary material (see  Sec.~\ref{DKT_model}).

\subsection{Machine Teaching} 
Estimating the representations used by humans is an important component for developing automated teaching algorithms and systems. 
Machine teaching algorithms address the teaching problem by generating sequences of instructional examples to show to novice learners in order to improve their ability on a given task~\cite{zhu2018overview}. 
Machine teaching has applications in crowdsourcing where the aim is to efficiently train crowdworkers, in addition to education where the goal is to train new experts, \eg in medical image analysis~\cite{cheng2020artificial,amiri2020training}. 

There is a growing body of work in computer vision that attempts to teach visual concepts to human learners, \eg~\cite{singla2014near,johns2015becoming,mac2018teaching,wang2021gradient,wang2021machine}.  
However, many of these works assume a fixed feature space that is generated before teaching begins \cite{singla2014near,johns2015becoming,mac2018teaching}. 
In one experiment, \cite{mac2018teaching} showed that representations that are better aligned with human perception result in improved learner performance on the downstream teaching task. 
While we do not explicitly investigate teaching algorithms in this work, we instead explore a setting where data is collected from humans engaged in learning a visual categorization task with instructional images selected by a `random' teacher. 
Importantly, the representations and learner parameterizations extracted by our model can be used directly with computer assisted teaching methods.

\section{Method}
Our goal is to estimate the image classification function used by a human learner that has been provided with a sequence of images and corresponding ground truth class labels as training data. 
We begin by outlining the problem, and then present our approach to human visual knowledge tracing.

\subsection{Problem Setup}
Given an image $\mathbf{x}$ as input, we model a human learner as a classification function that returns a response, $r^k = \argmax_c P(c | \mathbf{x}, \theta^k)$.  
Here, $r$ is a discrete class label representing the class response for learner $k$, \ie $r \in \{1, ... , C\}$, where $C$ is the number of possible classes, and $\theta^k$ are  unobserved parameters representing the state of the learner. 
Our learners are not stationary as their internal `knowledge state' changes depending on the information they have previously been exposed to that is relevant to the task.  
As a result, for a given learner $k$ we model their classification function at time $t$ as %
$r^k_{t+1} = \argmax_c~P(c | \mathbf{x}, \theta^k_t, \mathbf{x}^k_{1:t}, y^k_{1:t}, r^k_{1:t})$. 
Here, $(\mathbf{x}^k_{1:t}, y^k_{1:t}, r^k_{1:t})$ is the history of images, ground truth class labels, and responses that a learner $k$ has seen, and provided, up to and including time $t$.

Specifically, at each training time-step, a learner $k$ is presented with an image $\mathbf{x}$, they provide their response $r$, and are given feedback in the form of the correct class label $y$~(Fig. \ref{fig:data_summary}B). However, fitting a model for an individual learner with a single response per time-step is difficult. Alternatively, requesting more responses per time-step would reduce the number of teaching examples presented to the learner in the same amount of time. Instead, to overcome this limited information setting, we train a model $\phi$ across many learners, allowing the model to discover knowledge states and learning rules shared across all learners. 
Once trained, our model can make predictions for how a learner, who was not observed during training, will classify an image based on their prior classification behavior.

\subsection{Tracing Human Learners} 
Our tracing model $\phi$ can be decomposed into a feature extractor $f$, a classification function $\psi$, and a non-learned and non-linear transformation $\sigma$ (softmax). 
We explore how to represent the feature extractor and classification function.

A natural choice for the feature extractor $f$ is a Convolutional Neural Network (CNN). 
Given an image $\mathbf{x}$ as input, the feature extractor outputs a $D$ dimensional vector $\mathbf{z} = f(\mathbf{x})$. 
We will assume that all learners use the same underlying feature extractor which remains constant over time \ie $f = f^k_t$, and that they simply differ in the relative importance they place on different visual features.  
While these are both big assumptions to make, they are not overly restrictive. 
For example, a novice and an expert might engaged in the same visual classification task but differ in the set of visual features they select in order to make their decision. 
Furthermore, while we assume that the feature extractor remains constant over the time interval of our experiments, we do not assume that the classification function $\psi$ used by a learner remains static. 
In the next sections we will explore different choices for this classification function, comparing simple static classifiers with more expressive recurrent models.   

\subsubsection{Static Tracing Model.}  
The first model we explore is the simplest. 
Here we assume that all learners use the same classifier which does not vary over time.  
In this setting, $\psi$ is a multi-class linear classifier with a weight matrix $\mathbf{w}$ and per-class biases $\mathbf{b}$, 
\begin{equation}
\phi_{static}(\mathbf{x}) = \sigma(\psi(f(\mathbf{x}))) = \sigma(\mathbf{w}^\intercal f(\mathbf{x}) + \mathbf{b}).
\label{eqn:static}
\end{equation}
\noindent This model is similar to conventional metric learning approaches which do not attempt to capture any annotator specific differences related to individual biases or temporal changes. 
At training time we simply estimate one set of parameters for all learners. 
This model does not take the response history into account. 

\subsubsection{Time-Sensitive Tracing Model.}  
One obvious limitation of the static tracing model is that it does not take into account the fact that a learner will likely change over time, \ie they may be much worse at a new classification task early on, but may improve over time as they are shown sequences of example images along with their associated ground truth class labels. 
A more advanced model, that captures this temporal evolution, is one that has a different classifier for each time-step, 
\begin{equation}
\phi_{static\_time}(\mathbf{x}) = \sigma(\mathbf{w}_t^\intercal f(\mathbf{x}) + \mathbf{b}_t).
\label{eqn:static_time}
\end{equation}
Again, the same classifiers are shared across all learners, but in this case the weights and biases are different at each time-step, \ie $\mathbf{w}_t \neq \mathbf{w}_{t-1}$. 

\subsection{Recurrent Tracing Models} 
The previous tracing models do not account for the fact that individual learners may start with different levels of ability and update their internal knowledge state in different ways depending on the information that they are provided with. 
\cite{piech2015deep} showed that recurrent networks could be used to track the skill acquisition of human learners engaged in learning math quiz questions. 
Direct application of their model to our visual categorization setting is not possible as they assume one hot encodings of the query and learner responses as input. 
Their approach also uses large training sets -- on the order of thousands of learners and tens of thousands of interactions. 
Furthermore, they model knowledge acquisition at the `concept' (\ie visual category) and not `instance' (\ie a specific image) level, and thus their approach is not capable of making predictions for items not seen during training.  
We build on~\cite{piech2015deep} and adapt it to our visual category learning setting by presenting two different recurrent-based models for human visual knowledge tracing.

\subsubsection{Direct Response Model.}  
\label{dir_resp}
Our first model uses a recurrent network to directly predict the responses of a learner given their previous response history, 
\begin{equation}
\phi_{direct}(\mathbf{x}) = \sigma(\psi_{rnn}(\mathbf{z}^k_{1:t}, y^k_{1:t}, r^k_{1:t}, \mathbf{z}, y)).
\label{eqn:direct}
\end{equation}
Here, $\psi_{rnn}$ is a recurrent network (in practice we represent this using an LSTM~\cite{hochreiter1997long}) and $\mathbf{z}_t = f(\mathbf{x}_t)$ are visual features extracted from our CNN.

This model assumes that a learner's knowledge state at time $t$ is defined by the images they have previously seen and their past classification responses. Recurrent models can produce unique transformations for individual learners by conditioning on their hidden states. In this case, after the shared feature extractor transforms an image into a feature vector, the model modifies the feature vector with a series of non-linear transformations conditioned on the learner's hidden state. The final linear layer transforms the feature vector into a predicted response. 
Note that this model is also conditioned on the current query image $\mathbf{z} = f(\mathbf{x})$ and the corresponding ground truth class label  $y$. 

\subsubsection{Classifier Prediction Model.}  
\label{cls_pred}
\label{sec:class_pred}
Our second recurrent model attempts to provide a more interpretable approximation of human classification. 
In this case, instead of letting the recurrent model directly predict the probability for each response, it instead attempts to approximate the weights of a linear classifier used internally by the learner. 
Importantly, the values this classifier takes will depend on the response history of a given learner and will differ at each time-step, 

\begin{align}
 \mathbf{w}_t^k, \mathbf{b}_t^k &= \psi_{rnn}(\mathbf{z}^k_{1:t}, y^k_{1:t}, r^k_{1:t}, y), \label{eqn:class_pred_op}\\
 \phi_{cls\_pred}(\mathbf{x}) &= \sigma(\mathbf{w}_t^\intercal \mathbf{z} + \mathbf{b}_t). \label{eqn:cls_pred}
\end{align}

Unlike the previous direct prediction  recurrent model, we now explicitly represent the classification function used by an individual learner. 
Also, here the features $\mathbf{z}$ for the query image are not processed by the recurrent network in Eqn.~\ref{eqn:class_pred_op}.  
Instead they are evaluated using the much simpler predicted classifier weights in Eqn.~\ref{eqn:cls_pred}.  
This decoupling is advantageous in applications like machine teaching where we  have to query the tracing model multiple times at each time-step in order to determine the next image to show learners. 
Reducing the computation required to perform these queries will result in faster teacher algorithms.

\subsection{Training Tracing Models} 
We jointly estimate the parameters of the feature extractor $f$ and classification functions $\phi$ for each of the above models using a standard cross entropy loss, 
\begin{equation}
\mathcal{L} = -\frac{1}{K T} \sum_{k=1}^{K}\sum_{t=1}^{T}  \log(\phi(\mathbf{x}^k_{t})_r). 
\end{equation}
Here, $\phi(\mathbf{x}^k_{t})_r$ indicates the predicted probability from a model $\phi$ choosing class $r$, for learner $k$, at time-step $t$. 
The training objective aims to minimize the difference between the learner responses from our training set and the tracing model outputs.

\section{Experiments}
In this section we evaluate the different proposed models for visual knowledge tracing on data we collected from real human participants\footnote{Code and dataset - \href{https://github.com/nkondapa/VisualKnowledgeTracing}{https://github.com/nkondapa/VisualKnowledgeTracing}}.

\begin{figure}[t]
    \centering
    \includegraphics[width=0.9\textwidth]{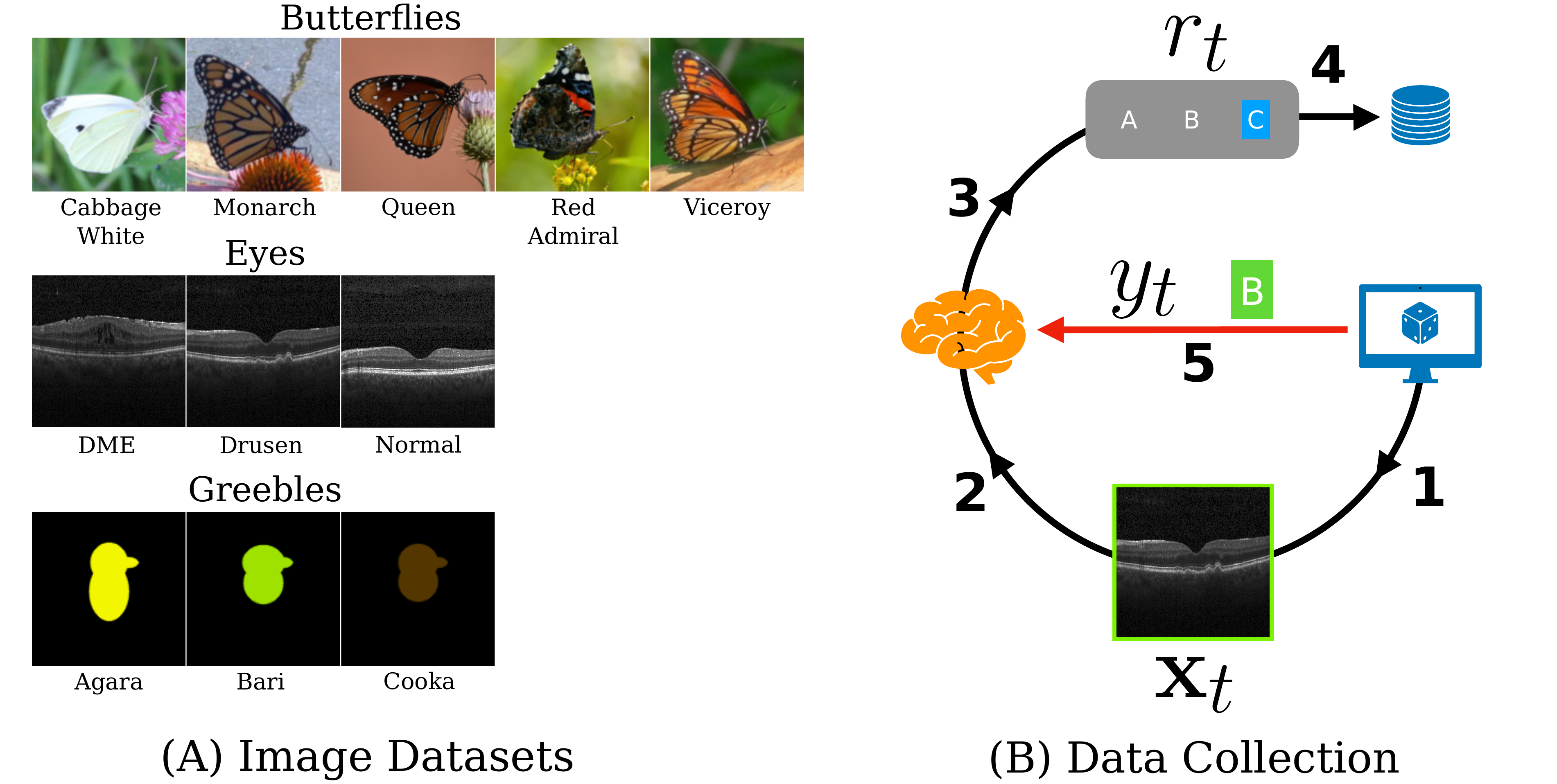}
    \caption{{\bf(A)} Example images from the three different datasets from our experiments. 
    `Butterflies' contains images of five different species and was originally presented in~\cite{mac2018teaching}. 
    `Eyes' contains optical coherence tomography images of the human retina from~\cite{kermany2018identifying}, and features two diseased classes and one normal one. 
    `Greebles' is a synthetic dataset we created where the three object classes vary in terms of shape and color.
    {\bf(B)} Data collection pipeline. A random image is selected (1), shown to the learner (2), and the learner provides a response (3). Their response is stored (4) and the correct class label is provided to them~(5). }
\label{fig:data_summary}
\end{figure}

\subsection{Datasets}
Traditional image classification datasets mostly contain labels produced by annotators familiar with the subject material, \eg~\cite{lin2014microsoft,van2015building,kuznetsova2020open}, or they have at least received detailed instructions and examples on how to annotate them, \eg~\cite{russakovsky2015imagenet}.  
As a result, these datasets do not contain annotations from learners engaged in learning a task and are thus not suitable for evaluating visual knowledge tracing. 
While some work has focused on teaching crowd learners (\eg ~\cite{singla2014near,mac2018teaching,wang2021gradient}), they often use teaching image sequences that are determined offline and fixed. 
For our tracing experiments, we require unbiased sequences of images that are randomly selected for each learner.  
Some of these existing works compare their approaches to a random image selection baseline, but the size of these random teaching subsets is insufficient for thorough evaluation of our different tracing approaches, \eg  \cite{mac2018teaching} have random selection data from only $\sim 40$ participants. 
Due to these limitations, we collected annotations from human learners for three challenging fine-grained visual classification datasets.

\subsubsection{Image Data.}   
We selected three different image datasets that cover three distinct domains: artificial data where we have full knowledge of the underlying distribution, medical image data, and images of different wildlife species. 
The first two datasets in particular are representative of the types of visual identification tasks that many humans are interested in learning. 

Our first dataset, `Butterflies', contains images from five different common species of North American butterflies. 
The `Cabbage White' class is immediately recognizable, 'Red Admiral' can be learned relatively easily, and the remaining three are difficult to discriminate. 
This dataset was originally used in \cite{mac2018teaching} and contains between 386 and 481 images per class, for a total of 2,224 images. 
Our second dataset, `Eyes', is a three-class subset of a large collection of publicly available images of the human retina from~\cite{kermany2018identifying}. 
It contains two diseased classes, `Diabetic Macular Edema' (DME) and `Drusen', and one `Normal' class.  
We manually selected 200 images from each class. 
The third dataset is a challenging synthetic one we created called `Greebles'.  
It contains three classes, where the underlying feature space used to generate the images is known by design. 
The relevant features are the body length and color, but the images also includes some irrelevant variation in the form of the head size and body width. 
The distinctions between the classes can be subtle, making the task challenging. 
It contains 1,200 images in total, with an even number per class. 
Visual examples for each of the datasets are presented in Fig.~\ref{fig:data_summary}A. 
Note that the single examples in Fig.~\ref{fig:data_summary}A do not covey the visual diversity of the datasets.

\subsubsection{Human Data Collection.} 
For each of the previously described datasets, we collected data from human participants that were engaged in learning the classification task. Each learner was presented with 30 training images and 15 test images. 
During the training phase they were provided with ground truth feedback indicating the correct class labels. 
This feedback was not provided in the test phase. 
For each image, we asked learners to rank the top three classes in order of most likely to least likely to be correct (see Sec.~\ref{label_ranking} for further information). 
The training and testing examples were randomly selected for each learner. 
An overview of the data collection process for one iteration, for one learner, is shown in Fig.~\ref{fig:data_summary}B.

The data was collected using a custom built a web application and the participants were recruited through the crowd sourcing app Prolific~\cite{prolific}. 
In total, we collected data from 150 learners for each dataset, where each individual could only do the task once. 
The median time spent on the Butterflies task, including training and testing, was $\sim11.7$ minutes with a median of $11$ correct on the test phase.  
The corresponding statistics for the Eyes dataset was $\sim12.1$ minutes with a median of $13$ correct, and for the Greebles dataset is was $\sim8.8$ minutes with a median of $9$ correct. 

\begin{figure}[t]
    \centering
    \includegraphics[trim={0pt 50pt -30pt 0pt},clip, width=1.0\textwidth]{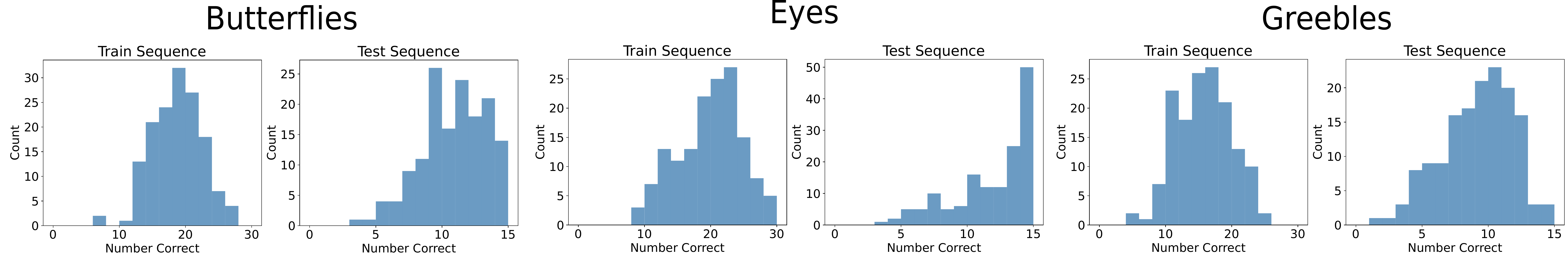}
    \caption{Human learner performance on our three datasets. For each dataset we provide histograms of learner performance on the respective training and testing sequences. 
    The training results are always worse as it include responses from all time-steps, including when the learner has just started the task and is unfamiliar with the classes. 
    For `Greebles', the worse performance on the test set compared to the other datasets indicates that learners find this task more challenging. 
    }
    \label{fig:hist_learner}
\end{figure}

Our human learners demonstrated learning across all three datasets. 
In Fig.~\ref{fig:hist_learner} we plot two histograms for each dataset, the first histogram displays the number of correct responses during the training (\ie teaching) phase and the second reports the same for the test phase. 
For all of the datasets, we see a right skew in the histogram of correct responses in the testing phase, indicating that most learners are providing correct answers after training. 
The skewness for each dataset is -0.45 for Butterflies, -0.81 for Eyes, and -0.37 for Greebles. 
The skewness is correlated to the amount of improvement learners showed on a given dataset. 
The numbers imply that the Greebles task is the most difficult and the learners demonstrate the least improvement. The Eyes dataset, on the other hand, is clearly the easiest. 
We provide additional analysis in the supplementary material~(Sec. {\ref{additional_results}}).

\subsection{Implementation Details}

All models we are considering primarily consist of a feature extractor and a classification function. 
The feature extractor is implemented as a CNN with eight layers (two convolutional, two max-pool, four linear), and is the same across all models (Sec.~\ref{cnn_details}). 
The feature extractor produces a 16 dimensional embedding for an input image which is then processed by the respective classification function. We evaluated one of our models on higher dimensional feature spaces, but found no impact on performance~(Sec. \ref{vary_embed_dim}). 
For the two recurrent models, $\phi_{direct}$ and $\phi_{cls\_pred}$, we use a three layer LSTM-based~\cite{hochreiter1997long} fully connected network with a hidden dimension of size 128. 
The output of the LSTM is passed to a small two-layer network that transforms the output into the desired representation (either a response or a classifier (see Sec. \ref{cls_pred})). We provide a detailed description of the architectures and inputs in the supplementary material~(Sec. \ref{RNN_Details}). 

All models are trained using mini-batch stochastic gradient descent with a batch size of 16 using the Adam optimizer~\cite{kingma2014adam}.   
For the feature extractor, we use a learning rate of 1e-5, and we use a learning rate of 1e-3 for each of the different classification functions. We train using a cross entropy loss.
Models are trained with early stopping. 
Training ends when the best validation loss does not improve for 35 epochs. 
The upper limit on the number of training epochs is 400. 
Data is split into train ($70\%$), validation ($13.3\%$), and test ($16.7\%$) splits at the learner level, \ie sequences from the same individual learner cannot be in more than one split.
To ensure more robust results, we validate models by re-shuffling the data five times and report the averages across the splits using micro and macro average precision.

\subsection{Results}

\subsubsection{Tracing Human Learners.}
We compute the precision and recall curves for each of the visual knowledge tracing models on the held-out learners and report the average precision for each. 
We also include one additional model for completeness, the ground truth baseline model (GT Label). 
This baseline does not fit any parameters and simply predicts the corresponding ground truth label of the image for all learners at all time-steps (instead of predicting the learner's response). 
Results are summarized in Table~\ref{tab:results}. 
Standard deviations are between 0.01 and 0.03 for all models, and do not change the interpretation of the results. 

We observe that the recurrent models, $\phi_{direct}$ and  $\phi_{cls\_pred}$, out-perform the baselines in tracing the learner on the Butterflies $({\sim}8\%)$ and Eyes datasets $({\sim}6\%)$. However, on the Greebles dataset, the static tracing model, $\phi_{static}$, outperforms the recurrent models by $({\sim}3\%)$. 
Both recurrent models are comparable in terms of performance, indicating the reduction in computation described in Sec.~\ref{sec:class_pred} does not come with a reduction in performance. The time-sensitive tracing model, $\phi_{static\_time}$,  is clearly the worst at tracing learners, likely owing to the unrealistic assumptions it makes about them.  

\begin{figure}[t]
    \centering
    \includegraphics[width=0.90\textwidth]{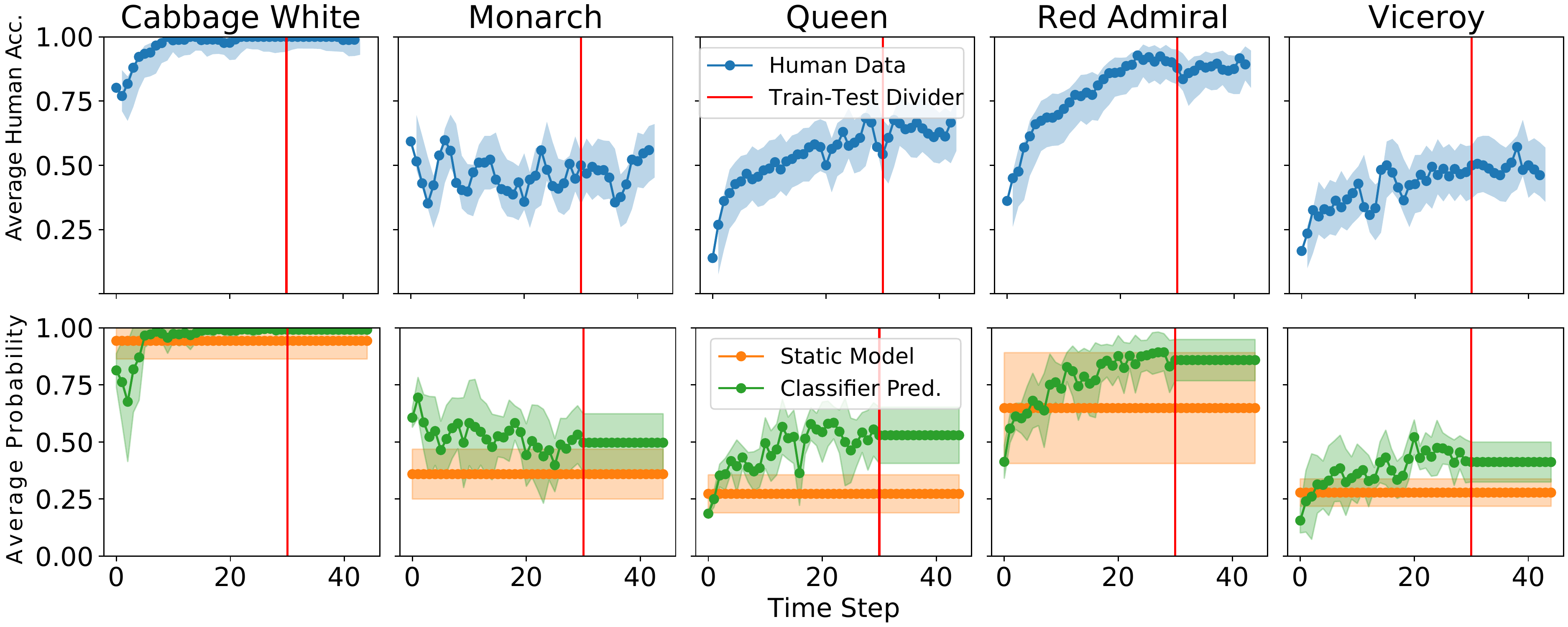}
    \caption{{\bf (Top)} The smoothed average human learner accuracy for class over time from the Butterflies dataset. The shadowed regions indicate confidence intervals as the number of samples in each time and class bin are not guaranteed to be the same.
    {\bf (Bottom)} The average probability of having a class correctly predicted by the static $\phi_{static}$ model (orange)  and the recurrent $\phi_{cls\_pred}$ model (green). 
    At each time-step, for each learner in the test set, the models predict class probabilities for $\sim 50$ images per class. 
    The probabilities are averaged (solid line) and the shadows indicate one standard deviation. In both rows, the red line indicates the point at which the learners switch from training to testing. After that, the models will continue to produce the same probabilities on the test images for the remaining time-steps as the sampled images do not change. 
    }
    \label{fig:trace_summary}
\end{figure}

Finally we explore the differences between how different models trace human learners. 
Fig.~\ref{fig:trace_summary} shows the average probability of predicting an image correctly for each time-step conditioned on each class for the Butterflies dataset. 
In the top row, we present the training and test-split average accuracy on each of the fives classes over time for the human learners. 
As previously noted, the training sequences contains 30 randomly selected images and the test sequence contains 15 images. 
We can see that on average, some classes are much easier than others. 
In the bottom row, we average model-predicted probabilities for ${\sim}50$ images in each class. To produce the probabilities for the recurrent $\phi_{cls\_pred}$ model, we processes the sequential data of the same set of learners in the top panel. 
The static tracing model, $\phi_{static}$, estimates a hyperplane for each class that roughly tracks the average probability of being correct per class. 
As expected, this model is not capable of capturing any learning behavior. 
In contrast, $\phi_{cls\_pred}$ more faithfully traces how the average probability evolves over time. 
Note, that we process each of the test images independently for $\phi_{cls\_pred}$.

\begin{table}[t]
\caption{Performance of different visual knowledge tracing approaches on data from human learners. 
We observe that our two recurrent based models, the direct response $\phi_{direct}$ and the classifier prediction $\phi_{cls\_pred}$, perform best on the Butterflies and Eyes dataset but are worse on the synthetic Greebles task. 
Learners found the Greebles task the most challenging, and as a result, there was much less learning occurring compared to the first two datasets. 
`GT Label' is an additional baseline that uses the corresponding ground truth class label $y$ as the prediction of the learner's response $r$. 
}
\centering
\resizebox{0.9\linewidth}{!}{

\begin{tabular}{l|llll|llll|llll|}
\cline{2-13}
 &
  \multicolumn{4}{c|}{\textbf{Greebles}} &
  \multicolumn{4}{c|}{\textbf{Eyes}} &
  \multicolumn{4}{c|}{\textbf{Butterflies}} \\ \cline{2-13} 
 &
  \multicolumn{2}{c|}{Train} &
  \multicolumn{2}{c|}{Test} &
  \multicolumn{2}{c|}{Train} &
  \multicolumn{2}{c|}{Test} &
  \multicolumn{2}{c|}{Train} &
  \multicolumn{2}{c|}{Test} \\ \cline{2-13} 
 &
  \multicolumn{1}{c|}{Micro} &
  \multicolumn{1}{c|}{Macro} &
  \multicolumn{1}{c|}{Micro} &
  \multicolumn{1}{c|}{Macro} &
  \multicolumn{1}{c|}{Micro} &
  \multicolumn{1}{c|}{Macro} &
  \multicolumn{1}{c|}{Micro} &
  \multicolumn{1}{c|}{Macro} &
  \multicolumn{1}{c|}{Micro} &
  \multicolumn{1}{c|}{Macro} &
  \multicolumn{1}{c|}{Micro} &
  \multicolumn{1}{c|}{Macro} \\ \hline
\multicolumn{1}{|l|}{GT Label} &
  \multicolumn{1}{l|}{0.48} &
  \multicolumn{1}{l|}{0.51} &
  \multicolumn{1}{l|}{0.58} &
  0.61 &
  \multicolumn{1}{l|}{0.56} &
  \multicolumn{1}{l|}{0.56} &
  \multicolumn{1}{l|}{0.69} &
  0.69 &
  \multicolumn{1}{l|}{0.45} &
  \multicolumn{1}{l|}{0.44} &
  \multicolumn{1}{l|}{0.50} &
  0.49 \\ \hline
\multicolumn{1}{|l|}{$\phi_{static}$} &
  \multicolumn{1}{l|}{0.63} &
  \multicolumn{1}{l|}{0.52} &
  \multicolumn{1}{l|}{0.67} &
  0.58 &
  \multicolumn{1}{l|}{0.60} &
  \multicolumn{1}{l|}{0.59} &
  \multicolumn{1}{l|}{0.67} &
  0.68 &
  \multicolumn{1}{l|}{\textbf{0.55}} &
  \multicolumn{1}{l|}{\textbf{0.53}} &
  \multicolumn{1}{l|}{\textbf{0.64}} &
  \textbf{0.61} \\ \hline
\multicolumn{1}{|l|}{$\phi_{static\_time}$} &
  \multicolumn{1}{l|}{0.52} &
  \multicolumn{1}{l|}{0.44} &
  \multicolumn{1}{l|}{0.49} &
  0.40 &
  \multicolumn{1}{l|}{0.34} &
  \multicolumn{1}{l|}{0.34} &
  \multicolumn{1}{l|}{0.33} &
  0.34 &
  \multicolumn{1}{l|}{0.54} &
  \multicolumn{1}{l|}{0.52} &
  \multicolumn{1}{l|}{0.61} &
  0.59 \\ \hline
\multicolumn{1}{|l|}{$\phi_{direct}$} &
  \multicolumn{1}{l|}{0.70} &
  \multicolumn{1}{l|}{0.59} &
  \multicolumn{1}{l|}{\textbf{0.77}} &
  0.64 &
  \multicolumn{1}{l|}{\textbf{0.66}} &
  \multicolumn{1}{l|}{\textbf{0.65}} &
  \multicolumn{1}{l|}{\textbf{0.75}} &
  \textbf{0.74} &
  \multicolumn{1}{l|}{\textbf{0.55}} &
  \multicolumn{1}{l|}{\textbf{0.53}} &
  \multicolumn{1}{l|}{0.60} &
  0.57 \\ \hline
\multicolumn{1}{|l|}{$\phi_{cls\_pred}$} &
  \multicolumn{1}{l|}{\textbf{0.71}} &
  \multicolumn{1}{l|}{\textbf{0.62}} &
  \multicolumn{1}{l|}{\textbf{0.77}} &
  \textbf{0.65} &
  \multicolumn{1}{l|}{0.65} &
  \multicolumn{1}{l|}{\textbf{0.65}} &
  \multicolumn{1}{l|}{0.74} &
  \textbf{0.74} &
  \multicolumn{1}{l|}{0.54} &
  \multicolumn{1}{l|}{0.52} &
  \multicolumn{1}{l|}{0.60} &
  0.57 \\ \hline
\end{tabular}

}
\label{tab:results}
\end{table}

\subsection{Discussion}

\subsubsection{Comparing Models.}

We observe that our recurrent models are quite effective at tracing human learner knowledge on visual categorization tasks for datasets where there is a clear learning signal. 
On the Greebles dataset, which is the most challenging and displays the least amount of learning, we see that the simple static tracing model $\phi_{static}$ is less prone to over-fitting and is thus marginally better. 
The time-sensitive tracing model $\phi_{static\_time}$ performs the worst overall. 
Unlike the recurrent methods, it is unable to share information between time-steps, forcing it to fit a classifier at each time-step with only ${\sim}90$ training points (the number of learners in the training set after we split out the validation and test sets). 
This makes it extremely prone to over-fitting on the limited training data that is available.

The direct response model $\phi_{direct}$ and the classifier prediction model $\phi_{cls\_pred}$ differ in their outputs and their inputs. 
$\phi_{direct}$ includes $\mathbf{z}$, the representation of the query image $\mathbf{x}$ to be classified by the learner, as input to the LSTM. 
$\phi_{cls\_pred}$ outputs weights of a multi-class linear classifier that is used to classify the representation $\mathbf{z}$. 
Unlike $\phi_{direct}$, $\mathbf{z}$ is not an input to the LSTM for $\phi_{cls\_pred}$. 
However, both models incorporate the ground truth label $y$ for $\mathbf{x}$ as input. 
These structural differences between the models make little difference to the tracing performance, but the $\phi_{cls\_pred}$ model is more efficient if it needs perform evaluation multiple times for new query items. 
In supplementary experiments~(\ref{RNN_variants}), we explore the impact of  removing the ground truth class label $y$ for the query image from the input. 
We observe that this results in a large decrease in performance for both models, suggesting that this class information is valuable to the model when making future predictions. 
Also in the supplement, we compare two cognitive models (prototype and exemplar~\cite{nosofsky2020contrasting})~(\ref{Cognitive Models}), a Transformer model~\cite{choi2020SAINT}~(\ref{transformer_method}), a pre-trained ResNet feature extractor~(\ref{ResNet_backbone_method}), and input meta-information to the tracing models~(\ref{PerClAcc}). We find that the cognitive, Transformer, and ResNet models do not out-perform the recurrent architectures, but are worth further exploration. Additionally, including meta-information in the input vectors results in a performance increase across all models we tested, suggesting this a promising direction for future work. Finally, we explore the  representations learned by the recurrent models (Secs. \ref{lstm_representation}, \ref{feature_space}).

\subsubsection{Limitations and Future Work.}
Currently, we train our feature encoder $f$ from scratch for each task on relatively small amounts of image  data. 
One source of improvement would be to pre-train the feature space so that it better reflects human visual similarity judgements.  
Such an improved feature space would provide a better starting point for task specific finetuning. 
Replacing the LSTM with an appropriately designed Transformer network~\cite{vaswani2017attention} is another change that  could result in greater flexibility for the model. 
Transformers are better able to capture long-term dependencies and have been shown to be useful in knowledge tracing on non-visual educational datasets~\cite{pandey2019self,choi2020SAINT}.
However, it remains to be seen if this would be valuable for visual knowledge tracing. 

Our approaches do not explicitly model memory decay -- the phenomenon of memory `fading' due to the passage of time~\cite{settles2016trainable,hunziker2019teaching}. 
This is likely to be more of a problem when tracing over longer time horizons, \eg days or weeks, as opposed to the multiple minute long sessions that our learners engage in. 
Similarly, given the short time durations of our teaching sessions, we assume that no significant `feature learning' is happening for individual learners. 
Instead we model learners as attending to a subset of the different visual features that are captured in our joint embedding space. 
In future work, it would be interesting to further explore if these two assumptions are valid.

\subsubsection{Applications.}
Successful tracing of human learners has implications for crowd-sourcing annotations, metric learning, and machine teaching. 
Early detection of poor annotators in crowd sourcing would reduce monetary and time costs in labelling datasets. 
Additionally, identifying annotators with `specialist' knowledge could allow for targeted crowd sourcing, tailoring to the abilities of each individual annotator. 
A successful tracing algorithm should be able to predict future performance with increasing confidence as the learners are being trained on the annotation task. %

The most impactful domain for a successful tracing algorithm is automated teaching, \eg teaching medical image interpretation skills~\cite{cheng2020artificial,amiri2020training}. 
Teaching humans is challenging because their knowledge state is unobserved, it changes over time, and the information they provide using current interfaces can be limited. 
In this work, we show that with a reasonable amount of training data (\ie data from $\sim{150}$  learners), and only a single response at each time-step, we are able to capture information about the learner's current knowledge for  visual classification tasks. 
The types of visual knowledge tracing approaches presented could be used in conjunction with machine teaching methods. 
The more a teaching algorithm knows about the learner, the more effective it can be when selecting examples to present to them.

\section{Conclusion}
More accurate models of human visual classification will lead to improved methods for crowd annotation collection, better techniques for automatically teaching visual knowledge to human learners, and perhaps provide us with insight into how we can build future artificial systems that are more data efficient. 
To this end, in this work we explored the problem of visual knowledge tracing -- the task of predicting the internal, potentially time varying, image classification function used by human learners. 
To do this, we presented a series of models that range in complexity from basic static linear classifiers all the way to recurrent models that take a learner's prior response history into account when making predictions about their future behavior. 
We collected new annotations for three challenging visual classification tasks from humans engaged in a visual learning task in order to benchmark the performance of these different models. 
Our results show that our recurrent neural network-based models resulted in the most faithful reproductions of unobserved learner predictions on real image datasets. 
Finally, we outlined limitations of our work and pointed to open questions that require further investigation. 
\small{
\par \noindent \textbf{Acknowledgements:} Thanks to the anonymous reviews for their valuable feedback. 
This work was in part supported by the Turing 2.0 `Enabling Advanced Autonomy' project funded by the EPSRC and the Alan Turing Institute and also by the Simons Collaboration on the Global Brain.
}

\clearpage
\appendix

\section{Additional Experiments}
\label{additional_experiments}
In this section, we present several additional models and also consider the impact of embedding dimension on performance. 

\subsection{RNN Variants}
\label{RNN_variants}
We can vary the input information to both of the recurrent models in three ways. The notation in parentheses maps to the entries in the later supplementary tables. \\ \\
(base): The models only receive the history of images, ground truth class labels, and learner responses 
\begin{equation}
\psi_{rnn}(\mathbf{z}^k_{1:t}, y^k_{1:t}, r^k_{1:t})).
\label{eqn:base}
\end{equation}
(y): In addition to the history, the model receives the ground truth class label of the image shown to the learner at the current time-step 
\begin{equation}
\psi_{rnn}(\mathbf{z}^k_{1:t}, y^k_{1:t}, r^k_{1:t}, y)).
\label{eqn:base+y}
\end{equation}
(y, $\mathbf{z}$): Finally, as in the main paper, the model can include both the ground truth class of the image and the representation of the image from the learned CNN 
\begin{equation}
\psi_{rnn}(\mathbf{z}^k_{1:t}, y^k_{1:t}, r^k_{1:t}, \mathbf{z}, y)).
\label{eqn:base+x+y}
\end{equation}

The results of the variants are presented in Tables~\ref{tab:butterflies_supp_results}, \ref{tab:eyes_supp_results}, and \ref{tab:greebles_supp_results}. 

\subsection{DKT}%
\label{DKT_model}
Next, we adapt Deep Knowledge Tracing (DKT)~\cite{piech2015deep} to our setting. 
We deviate from the original DKT method in two main ways. First, the types of queries (\eg math problems) in educational datasets do not allow for instance-level representations. Instead, skills (\ie question types) were jointly encoded with information about whether the problem was answered correctly by the learner. Second, the output of DKT was the learner's  probability of being correct for each skill, not a particular question instance. 

We modify the DKT algorithm to make it appropriate for the setting described in our work. We replace skills with the class-level label for an image and convert the output into a probability distribution over the class labels such that it can be trained with the cross-entropy loss

\begin{equation}
\phi_{dkt}(y) = \sigma(\psi_{rnn}(y^k_{1:t}, r^k_{1:t}, y))).
\label{eqn:dkt}
\end{equation}
At a high-level, this model variant encodes no instance-level (\ie image) information to make it's predictions. 

We observe that this DKT model ($\phi_{dkt}$) performs slightly worse in all cases, indicating that image information is valuable to enable the models to better trace learner performance. The results of the DKT variant are presented in Tables~\ref{tab:butterflies_supp_results}, \ref{tab:eyes_supp_results}, and \ref{tab:greebles_supp_results}. 

\subsection{Cognitive Models}%
\label{Cognitive Models}
Cognitive models make stronger assumptions on how humans learn. We modify the prototype and exemplar models described in the cognitive science literature~\cite{nosofsky2020contrasting} and evaluate them on our datasets.

\subsubsection{Prototype.}
The prototype model proposes that learners store a prototypical image for each class. Each new image is compared to the learner's class prototypes and the highest similarity class is selected. In our formulation, the class prototype is the average feature representation of previously seen images of that class. In the following equations, $\tau$ is the current time-step, $P^k_\tau(c)$ is the prototype of class c for learner k at time-step $\tau$, and $\delta$ is the dirac-delta function and acts as a selector for images from class c,

\begin{equation}
    P^k_\tau(c) = \frac{1}{\tau - 1} \cdot \sum^{\tau - 1}_t{z^k_t} * \delta(y^k_\tau - c), 
\end{equation}
\begin{equation}
    \hat{r}^k_{\tau}(c) = \frac{sim(P^k_\tau(c), z^k_\tau)}{\sum_{c}{sim(P^k_\tau(c), z^k_\tau)}}.
\end{equation}

\subsubsection{Exemplar.}
The exemplar model proposes that learners store previously seen images in a memory bank of exemplars. Query images are compared to all of the exemplars. The learner chooses the class with the highest total similarity to the query image. In the following equations, $E^k_\tau(c)$ is the sum of the class c similarity scores for learner $k$ at time-step $\tau$ with respect to the current image $z^k_\tau$. Following ~\cite{nosofsky2020contrasting}, we introduce a learnable parameter $\gamma$ to scale the similarities (this value is fixed to 1 in the prototype model),

\begin{equation}
    E^k_\tau(c) = \sum_t^{\tau - 1} {sim(z^k_t, z^k_\tau) \cdot \delta(y^k_\tau - c)},
\end{equation}
\begin{equation}
    \hat{r}^k_{\tau}(c) = \frac{sim(E^k_\tau(c), z^k_\tau)^\gamma}{\sum_{c}{sim(E^k_\tau(c), z^k_\tau)^\gamma}}.
\end{equation}

Both models compute similarity by using an exponential decay function over the Euclidean distance between feature representations of the images,

\begin{equation}
    sim(z_i, z_j) = e^{-c * d(z_i, z_j)}. 
\end{equation}

Finally, instead of learning the feature space separately with visual similarity experiments, we jointly estimate a CNN along with the model parameters to discover the feature space. 

We find that these models perform worse than the models presented in the main paper. However, simple modifications (like weighting the history of exemplars or images in the prototype) may help. Exploring the space of cognitive models is an interesting direction for future work. The results of these variants are presented in Tables~\ref{tab:butterflies_supp_results}, \ref{tab:eyes_supp_results}, and \ref{tab:greebles_supp_results}. 

\subsection{Transformers}%
\label{transformer_method}
Recently, the knowledge tracing community has found the Transformer architecture to be an effective model for tracing human learners in non-visual tasks. We modify the SAINT model~\cite{choi2020SAINT} for our visual learning setting. First, we introduce a CNN-based feature extraction stage to embed images. The encoder receives the current image's embedding and its ground truth label. The decoder gets the previous learner response. The decoder predicts the learner's response to the image (also passed to the encoder). 

The Transformer model does surprisingly poorly on these datasets. We expect that future work exploring Transformer architectures designed for this task will demonstrate performance on par with the recurrent models. The results of the Transformer model are presented in Tables~\ref{tab:butterflies_supp_results}, \ref{tab:eyes_supp_results}, and \ref{tab:greebles_supp_results}. 

\subsection{ResNet Backbone Experiments}
\label{ResNet_backbone_method}
We swap out our CNN backbone with a ResNet-18~\cite{he2016deep}  pre-trained on ImageNet. We freeze the weights in layers 1, 2 and 3, but leave layer 4 to be learned. The output of layer 4 is passed to a fully-connected layer that reduces the output of the layer to the desired dimensionality, as opposed to the final classifier used for the ImageNet classification task. 
The results of these experiments are presented in Table~\ref{tab:resnet_experiment}.

\subsection{Including Per-Class Accuracy as Input}
\label{PerClAcc}
We find that including some meta-information can help with tracing performance. 
To do this, we compute a learner's accuracy on each class at each time-step and concatenate this vector to the input of three tracing models ($\phi_{static}$, $\phi_{direct}$ and $\phi_{cls\_pred}$). We find a boost in performance across all models. The results are presented in Table~\ref{tab:per_class_acc_exp}, where we observe a boost in performance. 
It is likely that other sources of meta-information (such as time-taken on an example) will also help~\cite{choi2020SAINT+}.

\subsection{Varying Embedding Dimension}
\label{vary_embed_dim}
We demonstrate that varying the embedding dimension of the feature extractor has little effect on the performance of the direct response model (Table~\ref{tab:embedding_dimension}).

\begin{table}[]
\caption{Performance of all model variants on the Butterflies dataset. The model variant is denoted in the subscript corresponding to the same subscripts in~\ref{RNN_variants}. One can see that $\phi_{direct(base)}$ performs poorly for a recurrent model. This model does not have access to any information about the current time-step and is effectively guessing both the image that will be shown and the associated response. 
We also show the per-class average precision scores on the train sequence in addition to the micro and macro scores from before. These scores show that the benefit of the recurrent models appear primarily in classes that have large changes in average performance (\eg Red Admiral) over the training period. 
The models with $\dagger$ are models presented in Table 1 of the main paper. 
The scores are reported with their standard deviations and the top average performers in each column are in bold.
}
\centering
\resizebox{1.0\linewidth}{!}{

\begin{tabular}{
>{\columncolor[HTML]{FFFFFF}}c |
>{\columncolor[HTML]{FFFFFF}}l 
>{\columncolor[HTML]{FFFFFF}}l 
>{\columncolor[HTML]{FFFFFF}}l 
>{\columncolor[HTML]{FFFFFF}}l 
>{\columncolor[HTML]{FFFFFF}}l 
>{\columncolor[HTML]{FFFFFF}}l 
>{\columncolor[HTML]{FFFFFF}}l 
>{\columncolor[HTML]{FFFFFF}}l 
>{\columncolor[HTML]{FFFFFF}}l |}
\cline{2-10}
\multicolumn{1}{l|}{\cellcolor[HTML]{FFFFFF}\textbf{}} &
  \multicolumn{9}{c|}{\cellcolor[HTML]{FFFFFF}\textbf{Butterflies}} \\ \cline{2-10} 
\multicolumn{1}{l|}{\cellcolor[HTML]{FFFFFF}\textbf{}} &
  \multicolumn{7}{c|}{\cellcolor[HTML]{FFFFFF}Train} &
  \multicolumn{2}{c|}{\cellcolor[HTML]{FFFFFF}Test} \\ \cline{2-10} 
\multicolumn{1}{l|}{\cellcolor[HTML]{FFFFFF}\textbf{}} &
  \multicolumn{1}{c|}{\cellcolor[HTML]{FFFFFF}\begin{tabular}[c]{@{}c@{}}Cabbage\\ White\end{tabular}} &
  \multicolumn{1}{l|}{\cellcolor[HTML]{FFFFFF}Monarch} &
  \multicolumn{1}{l|}{\cellcolor[HTML]{FFFFFF}Queen} &
  \multicolumn{1}{c|}{\cellcolor[HTML]{FFFFFF}\begin{tabular}[c]{@{}c@{}}Red\\ Admiral\end{tabular}} &
  \multicolumn{1}{l|}{\cellcolor[HTML]{FFFFFF}Viceroy} &
  \multicolumn{1}{c|}{\cellcolor[HTML]{FFFFFF}Micro} &
  \multicolumn{1}{c|}{\cellcolor[HTML]{FFFFFF}Macro} &
  \multicolumn{1}{c|}{\cellcolor[HTML]{FFFFFF}Micro} &
  \multicolumn{1}{c|}{\cellcolor[HTML]{FFFFFF}Macro} \\ \hline
\multicolumn{1}{|c|}{\cellcolor[HTML]{FFFFFF}GT Label$\dagger$} &
  \multicolumn{1}{l|}{\cellcolor[HTML]{FFFFFF}0.94$\pm$0.03} &
  \multicolumn{1}{l|}{\cellcolor[HTML]{FFFFFF}0.32$\pm$0.01} &
  \multicolumn{1}{l|}{\cellcolor[HTML]{FFFFFF}0.36$\pm$0.04} &
  \multicolumn{1}{l|}{\cellcolor[HTML]{FFFFFF}0.65$\pm$0.04} &
  \multicolumn{1}{l|}{\cellcolor[HTML]{FFFFFF}0.27$\pm$0.01} &
  \multicolumn{1}{l|}{\cellcolor[HTML]{FFFFFF}0.48$\pm$0.02} &
  \multicolumn{1}{l|}{\cellcolor[HTML]{FFFFFF}0.51$\pm$0.02} &
  \multicolumn{1}{l|}{\cellcolor[HTML]{FFFFFF}0.58$\pm$0.02} &
  0.61$\pm$0.02 \\ \hline
\multicolumn{1}{|c|}{\cellcolor[HTML]{FFFFFF}$\phi_{static}\dagger$} &
  \multicolumn{1}{l|}{\cellcolor[HTML]{FFFFFF}0.95$\pm$0.01} &
  \multicolumn{1}{l|}{\cellcolor[HTML]{FFFFFF}0.37$\pm$0.02} &
  \multicolumn{1}{l|}{\cellcolor[HTML]{FFFFFF}0.34$\pm$0.06} &
  \multicolumn{1}{l|}{\cellcolor[HTML]{FFFFFF}0.66$\pm$0.03} &
  \multicolumn{1}{l|}{\cellcolor[HTML]{FFFFFF}0.27$\pm$0.03} &
  \multicolumn{1}{l|}{\cellcolor[HTML]{FFFFFF}0.63$\pm$0.02} &
  \multicolumn{1}{l|}{\cellcolor[HTML]{FFFFFF}0.52$\pm$0.02} &
  \multicolumn{1}{l|}{\cellcolor[HTML]{FFFFFF}0.67$\pm$0.02} &
  0.58$\pm$0.03 \\ \hline
\multicolumn{1}{|c|}{\cellcolor[HTML]{FFFFFF}$\phi_{static\_time}\dagger$} &
  \multicolumn{1}{l|}{\cellcolor[HTML]{FFFFFF}0.97$\pm$0.01} &
  \multicolumn{1}{l|}{\cellcolor[HTML]{FFFFFF}0.34$\pm$0.03} &
  \multicolumn{1}{l|}{\cellcolor[HTML]{FFFFFF}0.29$\pm$0.03} &
  \multicolumn{1}{l|}{\cellcolor[HTML]{FFFFFF}0.35$\pm$0.03} &
  \multicolumn{1}{l|}{\cellcolor[HTML]{FFFFFF}0.24$\pm$0.02} &
  \multicolumn{1}{l|}{\cellcolor[HTML]{FFFFFF}0.52$\pm$0.02} &
  \multicolumn{1}{l|}{\cellcolor[HTML]{FFFFFF}0.44$\pm$0.01} &
  \multicolumn{1}{l|}{\cellcolor[HTML]{FFFFFF}0.49$\pm$0.01} &
  0.40$\pm$0.01 \\ \hline
\multicolumn{1}{|c|}{\cellcolor[HTML]{FFFFFF}$\phi_{dkt}$} &
  \multicolumn{1}{l|}{\cellcolor[HTML]{FFFFFF}0.95$\pm$0.02} &
  \multicolumn{1}{l|}{\cellcolor[HTML]{FFFFFF}0.43$\pm$0.02} &
  \multicolumn{1}{l|}{\cellcolor[HTML]{FFFFFF}0.45$\pm$0.04} &
  \multicolumn{1}{l|}{\cellcolor[HTML]{FFFFFF}0.72$\pm$0.05} &
  \multicolumn{1}{l|}{\cellcolor[HTML]{FFFFFF}0.33$\pm$0.06} &
  \multicolumn{1}{l|}{\cellcolor[HTML]{FFFFFF}0.67$\pm$0.02} &
  \multicolumn{1}{l|}{\cellcolor[HTML]{FFFFFF}0.57$\pm$0.02} &
  \multicolumn{1}{l|}{\cellcolor[HTML]{FFFFFF}0.74$\pm$0.02} &
  0.64$\pm$0.02 \\ \hline
\multicolumn{1}{|c|}{\cellcolor[HTML]{FFFFFF}$\phi_{transformer}$} &
  \multicolumn{1}{l|}{\cellcolor[HTML]{FFFFFF}0.96$\pm$0.01} &
  \multicolumn{1}{l|}{\cellcolor[HTML]{FFFFFF}0.36$\pm$0.03} &
  \multicolumn{1}{l|}{\cellcolor[HTML]{FFFFFF}0.34$\pm$0.02} &
  \multicolumn{1}{l|}{\cellcolor[HTML]{FFFFFF}0.69$\pm$0.06} &
  \multicolumn{1}{l|}{\cellcolor[HTML]{FFFFFF}0.26$\pm$0.01} &
  \multicolumn{1}{l|}{\cellcolor[HTML]{FFFFFF}0.62$\pm$0.02} &
  \multicolumn{1}{l|}{\cellcolor[HTML]{FFFFFF}0.52$\pm$0.02} &
  \multicolumn{1}{l|}{\cellcolor[HTML]{FFFFFF}0.68$\pm$0.05} &
  0.58$\pm$0.04 \\ \hline
\multicolumn{1}{|c|}{\cellcolor[HTML]{FFFFFF}$\phi_{prototype}$} &
  \multicolumn{1}{l|}{\cellcolor[HTML]{FFFFFF}0.95$\pm$0.02} &
  \multicolumn{1}{l|}{\cellcolor[HTML]{FFFFFF}0.32$\pm$0.03} &
  \multicolumn{1}{l|}{\cellcolor[HTML]{FFFFFF}0.30$\pm$0.05} &
  \multicolumn{1}{l|}{\cellcolor[HTML]{FFFFFF}0.59$\pm$0.08} &
  \multicolumn{1}{l|}{\cellcolor[HTML]{FFFFFF}0.27$\pm$0.03} &
  \multicolumn{1}{l|}{\cellcolor[HTML]{FFFFFF}0.53$\pm$0.03} &
  \multicolumn{1}{l|}{\cellcolor[HTML]{FFFFFF}0.48$\pm$0.03} &
  \multicolumn{1}{l|}{\cellcolor[HTML]{FFFFFF}0.63$\pm$0.01} &
  0.54$\pm$0.02 \\ \hline
\multicolumn{1}{|c|}{\cellcolor[HTML]{FFFFFF}$\phi_{exemplar}$} &
  \multicolumn{1}{l|}{\cellcolor[HTML]{FFFFFF}0.90$\pm$0.03} &
  \multicolumn{1}{l|}{\cellcolor[HTML]{FFFFFF}0.30$\pm$0.03} &
  \multicolumn{1}{l|}{\cellcolor[HTML]{FFFFFF}0.27$\pm$0.04} &
  \multicolumn{1}{l|}{\cellcolor[HTML]{FFFFFF}0.35$\pm$0.13} &
  \multicolumn{1}{l|}{\cellcolor[HTML]{FFFFFF}0.26$\pm$0.03} &
  \multicolumn{1}{l|}{\cellcolor[HTML]{FFFFFF}0.44$\pm$0.05} &
  \multicolumn{1}{l|}{\cellcolor[HTML]{FFFFFF}0.42$\pm$0.04} &
  \multicolumn{1}{l|}{\cellcolor[HTML]{FFFFFF}0.53$\pm$0.07} &
  0.45$\pm$0.07 \\ \hline
\multicolumn{1}{|c|}{\cellcolor[HTML]{FFFFFF}$\phi_{direct(base)}$} &
  \multicolumn{1}{l|}{\cellcolor[HTML]{FFFFFF}0.34$\pm$0.02} &
  \multicolumn{1}{l|}{\cellcolor[HTML]{FFFFFF}0.29$\pm$0.03} &
  \multicolumn{1}{l|}{\cellcolor[HTML]{FFFFFF}0.23$\pm$0.02} &
  \multicolumn{1}{l|}{\cellcolor[HTML]{FFFFFF}0.26$\pm$0.02} &
  \multicolumn{1}{l|}{\cellcolor[HTML]{FFFFFF}0.22$\pm$0.03} &
  \multicolumn{1}{l|}{\cellcolor[HTML]{FFFFFF}0.27$\pm$0.01} &
  \multicolumn{1}{l|}{\cellcolor[HTML]{FFFFFF}0.27$\pm$0.01} &
  \multicolumn{1}{l|}{\cellcolor[HTML]{FFFFFF}0.20$\pm$0.01} &
  0.20$\pm$0.01 \\ \hline
\multicolumn{1}{|c|}{\cellcolor[HTML]{FFFFFF}$\phi_{direct(y)}$} &
  \multicolumn{1}{l|}{\cellcolor[HTML]{FFFFFF}0.97$\pm$0.02} &
  \multicolumn{1}{l|}{\cellcolor[HTML]{FFFFFF}0.43$\pm$0.03} &
  \multicolumn{1}{l|}{\cellcolor[HTML]{FFFFFF}\textbf{0.48$\pm$0.05}} &
  \multicolumn{1}{l|}{\cellcolor[HTML]{FFFFFF}0.76$\pm$0.07} &
  \multicolumn{1}{l|}{\cellcolor[HTML]{FFFFFF}\textbf{0.38$\pm$0.04}} &
  \multicolumn{1}{l|}{\cellcolor[HTML]{FFFFFF}\textbf{0.71$\pm$0.02}} &
  \multicolumn{1}{l|}{\cellcolor[HTML]{FFFFFF}0.60$\pm$0.02} &
  \multicolumn{1}{l|}{\cellcolor[HTML]{FFFFFF}\textbf{0.78$\pm$0.02}} &
  \textbf{0.66$\pm$0.01} \\ \hline
\multicolumn{1}{|c|}{\cellcolor[HTML]{FFFFFF}$\phi_{direct(y, \mathbf{z})}\dagger$} &
  \multicolumn{1}{l|}{\cellcolor[HTML]{FFFFFF}0.97$\pm$0.01} &
  \multicolumn{1}{l|}{\cellcolor[HTML]{FFFFFF}0.41$\pm$0.04} &
  \multicolumn{1}{l|}{\cellcolor[HTML]{FFFFFF}0.44$\pm$0.06} &
  \multicolumn{1}{l|}{\cellcolor[HTML]{FFFFFF}0.77$\pm$0.07} &
  \multicolumn{1}{l|}{\cellcolor[HTML]{FFFFFF}0.36$\pm$0.03} &
  \multicolumn{1}{l|}{\cellcolor[HTML]{FFFFFF}0.70$\pm$0.03} &
  \multicolumn{1}{l|}{\cellcolor[HTML]{FFFFFF}0.59$\pm$0.02} &
  \multicolumn{1}{l|}{\cellcolor[HTML]{FFFFFF}0.77$\pm$0.03} &
  0.64$\pm$0.03 \\ \hline
\multicolumn{1}{|c|}{\cellcolor[HTML]{FFFFFF}$\phi_{cls\_pred(base)}$} &
  \multicolumn{1}{l|}{\cellcolor[HTML]{FFFFFF}0.96$\pm$0.01} &
  \multicolumn{1}{l|}{\cellcolor[HTML]{FFFFFF}0.41$\pm$0.06} &
  \multicolumn{1}{l|}{\cellcolor[HTML]{FFFFFF}0.32$\pm$0.05} &
  \multicolumn{1}{l|}{\cellcolor[HTML]{FFFFFF}0.59$\pm$0.14} &
  \multicolumn{1}{l|}{\cellcolor[HTML]{FFFFFF}0.25$\pm$0.02} &
  \multicolumn{1}{l|}{\cellcolor[HTML]{FFFFFF}0.59$\pm$0.06} &
  \multicolumn{1}{l|}{\cellcolor[HTML]{FFFFFF}0.51$\pm$0.05} &
  \multicolumn{1}{l|}{\cellcolor[HTML]{FFFFFF}0.61$\pm$0.06} &
  0.52$\pm$0.05 \\ \hline
\multicolumn{1}{|c|}{\cellcolor[HTML]{FFFFFF}$\phi_{cls\_pred(y)}\dagger$} &
  \multicolumn{1}{l|}{\cellcolor[HTML]{FFFFFF}\textbf{0.98$\pm$0.01}} &
  \multicolumn{1}{l|}{\cellcolor[HTML]{FFFFFF}\textbf{0.46$\pm$0.02}} &
  \multicolumn{1}{l|}{\cellcolor[HTML]{FFFFFF}\textbf{0.48$\pm$0.04}} &
  \multicolumn{1}{l|}{\cellcolor[HTML]{FFFFFF}\textbf{0.78$\pm$0.05}} &
  \multicolumn{1}{l|}{\cellcolor[HTML]{FFFFFF}\textbf{0.38$\pm$0.02}} &
  \multicolumn{1}{l|}{\cellcolor[HTML]{FFFFFF}\textbf{0.71$\pm$0.02}} &
  \multicolumn{1}{l|}{\cellcolor[HTML]{FFFFFF}\textbf{0.62$\pm$0.01}} &
  \multicolumn{1}{l|}{\cellcolor[HTML]{FFFFFF}0.77$\pm$0.02} &
  0.65$\pm$0.02 \\ \hline
\multicolumn{1}{|c|}{\cellcolor[HTML]{FFFFFF}$\phi_{cls\_pred(y, \mathbf{z})}$} &
  \multicolumn{1}{l|}{\cellcolor[HTML]{FFFFFF}\textbf{0.98$\pm$0.01}} &
  \multicolumn{1}{l|}{\cellcolor[HTML]{FFFFFF}0.45$\pm$0.01} &
  \multicolumn{1}{l|}{\cellcolor[HTML]{FFFFFF}0.47$\pm$0.04} &
  \multicolumn{1}{l|}{\cellcolor[HTML]{FFFFFF}\textbf{0.78$\pm$0.05}} &
  \multicolumn{1}{l|}{\cellcolor[HTML]{FFFFFF}0.37$\pm$0.02} &
  \multicolumn{1}{l|}{\cellcolor[HTML]{FFFFFF}0.70$\pm$0.02} &
  \multicolumn{1}{l|}{\cellcolor[HTML]{FFFFFF}0.61$\pm$0.01} &
  \multicolumn{1}{l|}{\cellcolor[HTML]{FFFFFF}0.77$\pm$0.03} &
  \textbf{0.66$\pm$0.02} \\ \hline
\end{tabular}
}
\label{tab:butterflies_supp_results}
\end{table}

\begin{table}[h]
\caption{Performance of all model variants on the Eyes dataset. Please see the caption of Table~\ref{tab:butterflies_supp_results} for more details. 
}
\centering
\resizebox{1.0\linewidth}{!}{

\begin{tabular}{
>{\columncolor[HTML]{FFFFFF}}c |
>{\columncolor[HTML]{FFFFFF}}l 
>{\columncolor[HTML]{FFFFFF}}l 
>{\columncolor[HTML]{FFFFFF}}l 
>{\columncolor[HTML]{FFFFFF}}l 
>{\columncolor[HTML]{FFFFFF}}l 
>{\columncolor[HTML]{FFFFFF}}l 
>{\columncolor[HTML]{FFFFFF}}l |}
\cline{2-8}
\multicolumn{1}{l|}{\cellcolor[HTML]{FFFFFF}} &
  \multicolumn{7}{c|}{\cellcolor[HTML]{FFFFFF}\textbf{Eyes}} \\ \cline{2-8} 
\multicolumn{1}{l|}{\cellcolor[HTML]{FFFFFF}} &
  \multicolumn{5}{c|}{\cellcolor[HTML]{FFFFFF}Train} &
  \multicolumn{2}{c|}{\cellcolor[HTML]{FFFFFF}Test} \\ \cline{2-8} 
\multicolumn{1}{l|}{\cellcolor[HTML]{FFFFFF}} &
  \multicolumn{1}{c|}{\cellcolor[HTML]{FFFFFF}DME} &
  \multicolumn{1}{l|}{\cellcolor[HTML]{FFFFFF}Drusen} &
  \multicolumn{1}{l|}{\cellcolor[HTML]{FFFFFF}Normal} &
  \multicolumn{1}{c|}{\cellcolor[HTML]{FFFFFF}Micro} &
  \multicolumn{1}{c|}{\cellcolor[HTML]{FFFFFF}Macro} &
  \multicolumn{1}{c|}{\cellcolor[HTML]{FFFFFF}Micro} &
  \multicolumn{1}{c|}{\cellcolor[HTML]{FFFFFF}Macro} \\ \hline
\multicolumn{1}{|c|}{\cellcolor[HTML]{FFFFFF}GT Label$\dagger$} &
  \multicolumn{1}{l|}{\cellcolor[HTML]{FFFFFF}0.56$\pm$0.02} &
  \multicolumn{1}{l|}{\cellcolor[HTML]{FFFFFF}0.54$\pm$0.03} &
  \multicolumn{1}{l|}{\cellcolor[HTML]{FFFFFF}0.58$\pm$0.02} &
  \multicolumn{1}{l|}{\cellcolor[HTML]{FFFFFF}0.56$\pm$0.02} &
  \multicolumn{1}{l|}{\cellcolor[HTML]{FFFFFF}0.56$\pm$0.02} &
  \multicolumn{1}{l|}{\cellcolor[HTML]{FFFFFF}0.69$\pm$0.02} &
  0.69$\pm$0.01 \\ \hline
\multicolumn{1}{|c|}{\cellcolor[HTML]{FFFFFF}$\phi_{static}\dagger$} &
  \multicolumn{1}{l|}{\cellcolor[HTML]{FFFFFF}0.63$\pm$0.03} &
  \multicolumn{1}{l|}{\cellcolor[HTML]{FFFFFF}0.53$\pm$0.05} &
  \multicolumn{1}{l|}{\cellcolor[HTML]{FFFFFF}0.62$\pm$0.02} &
  \multicolumn{1}{l|}{\cellcolor[HTML]{FFFFFF}0.60$\pm$0.03} &
  \multicolumn{1}{l|}{\cellcolor[HTML]{FFFFFF}0.59$\pm$0.03} &
  \multicolumn{1}{l|}{\cellcolor[HTML]{FFFFFF}0.67$\pm$0.02} &
  0.68$\pm$0.02 \\ \hline
\multicolumn{1}{|c|}{\cellcolor[HTML]{FFFFFF}$\phi_{static_time}\dagger$} &
  \multicolumn{1}{l|}{\cellcolor[HTML]{FFFFFF}0.32$\pm$0.01} &
  \multicolumn{1}{l|}{\cellcolor[HTML]{FFFFFF}0.35$\pm$0.01} &
  \multicolumn{1}{l|}{\cellcolor[HTML]{FFFFFF}0.36$\pm$0.01} &
  \multicolumn{1}{l|}{\cellcolor[HTML]{FFFFFF}0.34$\pm$0.00} &
  \multicolumn{1}{l|}{\cellcolor[HTML]{FFFFFF}0.34$\pm$0.01} &
  \multicolumn{1}{l|}{\cellcolor[HTML]{FFFFFF}0.33$\pm$0.01} &
  0.34$\pm$0.02 \\ \hline
\multicolumn{1}{|c|}{\cellcolor[HTML]{FFFFFF}$\phi_{dkt}$} &
  \multicolumn{1}{l|}{\cellcolor[HTML]{FFFFFF}0.63$\pm$0.02} &
  \multicolumn{1}{l|}{\cellcolor[HTML]{FFFFFF}0.60$\pm$0.03} &
  \multicolumn{1}{l|}{\cellcolor[HTML]{FFFFFF}0.65$\pm$0.02} &
  \multicolumn{1}{l|}{\cellcolor[HTML]{FFFFFF}0.63$\pm$0.02} &
  \multicolumn{1}{l|}{\cellcolor[HTML]{FFFFFF}0.63$\pm$0.02} &
  \multicolumn{1}{l|}{\cellcolor[HTML]{FFFFFF}0.74$\pm$0.03} &
  0.73$\pm$0.03 \\ \hline
\multicolumn{1}{|c|}{\cellcolor[HTML]{FFFFFF}$\phi_{transformer}$} &
  \multicolumn{1}{l|}{\cellcolor[HTML]{FFFFFF}0.41$\pm$0.09} &
  \multicolumn{1}{l|}{\cellcolor[HTML]{FFFFFF}0.41$\pm$0.05} &
  \multicolumn{1}{l|}{\cellcolor[HTML]{FFFFFF}0.42$\pm$0.09} &
  \multicolumn{1}{l|}{\cellcolor[HTML]{FFFFFF}0.41$\pm$0.08} &
  \multicolumn{1}{l|}{\cellcolor[HTML]{FFFFFF}0.41$\pm$0.08} &
  \multicolumn{1}{l|}{\cellcolor[HTML]{FFFFFF}0.37$\pm$0.02} &
  0.39$\pm$0.02 \\ \hline
\multicolumn{1}{|c|}{\cellcolor[HTML]{FFFFFF}$\phi_{prototype}$} &
  \multicolumn{1}{l|}{\cellcolor[HTML]{FFFFFF}0.61$\pm$0.04} &
  \multicolumn{1}{l|}{\cellcolor[HTML]{FFFFFF}0.50$\pm$0.04} &
  \multicolumn{1}{l|}{\cellcolor[HTML]{FFFFFF}0.56$\pm$0.04} &
  \multicolumn{1}{l|}{\cellcolor[HTML]{FFFFFF}0.56$\pm$0.03} &
  \multicolumn{1}{l|}{\cellcolor[HTML]{FFFFFF}0.56$\pm$0.03} &
  \multicolumn{1}{l|}{\cellcolor[HTML]{FFFFFF}0.65$\pm$0.04} &
  0.65$\pm$0.05 \\ \hline
\multicolumn{1}{|c|}{\cellcolor[HTML]{FFFFFF}$\phi_{exemplar}$} &
  \multicolumn{1}{l|}{\cellcolor[HTML]{FFFFFF}0.57$\pm$0.02} &
  \multicolumn{1}{l|}{\cellcolor[HTML]{FFFFFF}0.48$\pm$0.02} &
  \multicolumn{1}{l|}{\cellcolor[HTML]{FFFFFF}0.59$\pm$0.03} &
  \multicolumn{1}{l|}{\cellcolor[HTML]{FFFFFF}0.54$\pm$0.02} &
  \multicolumn{1}{l|}{\cellcolor[HTML]{FFFFFF}0.55$\pm$0.02} &
  \multicolumn{1}{l|}{\cellcolor[HTML]{FFFFFF}0.68$\pm$0.04} &
  0.67$\pm$0.04 \\ \hline
\multicolumn{1}{|c|}{\cellcolor[HTML]{FFFFFF}$\phi_{direct(base)}$} &
  \multicolumn{1}{l|}{\cellcolor[HTML]{FFFFFF}0.38$\pm$0.02} &
  \multicolumn{1}{l|}{\cellcolor[HTML]{FFFFFF}0.40$\pm$0.01} &
  \multicolumn{1}{l|}{\cellcolor[HTML]{FFFFFF}0.38$\pm$0.02} &
  \multicolumn{1}{l|}{\cellcolor[HTML]{FFFFFF}0.38$\pm$0.01} &
  \multicolumn{1}{l|}{\cellcolor[HTML]{FFFFFF}0.39$\pm$0.01} &
  \multicolumn{1}{l|}{\cellcolor[HTML]{FFFFFF}0.35$\pm$0.01} &
  0.34$\pm$0.01 \\ \hline
\multicolumn{1}{|c|}{\cellcolor[HTML]{FFFFFF}$\phi_{direct(y)}$} &
  \multicolumn{1}{l|}{\cellcolor[HTML]{FFFFFF}\textbf{0.65$\pm$0.03}} &
  \multicolumn{1}{l|}{\cellcolor[HTML]{FFFFFF}\textbf{0.62$\pm$0.03}} &
  \multicolumn{1}{l|}{\cellcolor[HTML]{FFFFFF}0.68$\pm$0.03} &
  \multicolumn{1}{l|}{\cellcolor[HTML]{FFFFFF}\textbf{0.66$\pm$0.02}} &
  \multicolumn{1}{l|}{\cellcolor[HTML]{FFFFFF}\textbf{0.65$\pm$0.02}} &
  \multicolumn{1}{l|}{\cellcolor[HTML]{FFFFFF}\textbf{0.75$\pm$0.01}} &
  0.73$\pm$0.02 \\ \hline
\multicolumn{1}{|c|}{\cellcolor[HTML]{FFFFFF}$\phi_{direct(y, \mathbf{z})}\dagger$} &
  \multicolumn{1}{l|}{\cellcolor[HTML]{FFFFFF}0.64$\pm$0.02} &
  \multicolumn{1}{l|}{\cellcolor[HTML]{FFFFFF}\textbf{0.62$\pm$0.03}} &
  \multicolumn{1}{l|}{\cellcolor[HTML]{FFFFFF}\textbf{0.69$\pm$0.02}} &
  \multicolumn{1}{l|}{\cellcolor[HTML]{FFFFFF}\textbf{0.66$\pm$0.02}} &
  \multicolumn{1}{l|}{\cellcolor[HTML]{FFFFFF}\textbf{0.65$\pm$0.02}} &
  \multicolumn{1}{l|}{\cellcolor[HTML]{FFFFFF}\textbf{0.75$\pm$0.01}} &
  \textbf{0.74$\pm$0.02} \\ \hline
\multicolumn{1}{|c|}{\cellcolor[HTML]{FFFFFF}$\phi_{cls\_pred(base)}$} &
  \multicolumn{1}{l|}{\cellcolor[HTML]{FFFFFF}0.48$\pm$0.05} &
  \multicolumn{1}{l|}{\cellcolor[HTML]{FFFFFF}0.39$\pm$0.02} &
  \multicolumn{1}{l|}{\cellcolor[HTML]{FFFFFF}0.48$\pm$0.05} &
  \multicolumn{1}{l|}{\cellcolor[HTML]{FFFFFF}0.45$\pm$0.04} &
  \multicolumn{1}{l|}{\cellcolor[HTML]{FFFFFF}0.45$\pm$0.03} &
  \multicolumn{1}{l|}{\cellcolor[HTML]{FFFFFF}0.44$\pm$0.02} &
  0.44$\pm$0.02 \\ \hline
\multicolumn{1}{|c|}{\cellcolor[HTML]{FFFFFF}$\phi_{cls\_pred(y)}\dagger$} &
  \multicolumn{1}{l|}{\cellcolor[HTML]{FFFFFF}\textbf{0.65$\pm$0.02}} &
  \multicolumn{1}{l|}{\cellcolor[HTML]{FFFFFF}\textbf{0.62$\pm$0.03}} &
  \multicolumn{1}{l|}{\cellcolor[HTML]{FFFFFF}\textbf{0.69$\pm$0.01}} &
  \multicolumn{1}{l|}{\cellcolor[HTML]{FFFFFF}0.65$\pm$0.03} &
  \multicolumn{1}{l|}{\cellcolor[HTML]{FFFFFF}\textbf{0.65$\pm$0.02}} &
  \multicolumn{1}{l|}{\cellcolor[HTML]{FFFFFF}0.74$\pm$0.02} &
  \textbf{0.74$\pm$0.04} \\ \hline
\multicolumn{1}{|c|}{\cellcolor[HTML]{FFFFFF}$\phi_{cls\_pred(y, \mathbf{z})}$} &
  \multicolumn{1}{l|}{\cellcolor[HTML]{FFFFFF}0.64$\pm$0.01} &
  \multicolumn{1}{l|}{\cellcolor[HTML]{FFFFFF}\textbf{0.62$\pm$0.03}} &
  \multicolumn{1}{l|}{\cellcolor[HTML]{FFFFFF}\textbf{0.69$\pm$0.02}} &
  \multicolumn{1}{l|}{\cellcolor[HTML]{FFFFFF}0.65$\pm$0.02} &
  \multicolumn{1}{l|}{\cellcolor[HTML]{FFFFFF}\textbf{0.65$\pm$0.02}} &
  \multicolumn{1}{l|}{\cellcolor[HTML]{FFFFFF}\textbf{0.75$\pm$0.01}} &
  \textbf{0.74$\pm$0.02} \\ \hline
\end{tabular}
}
\label{tab:eyes_supp_results}
\end{table}

\begin{table}[h]
\caption{Performance of all model variants on the Greebles dataset. 
Please see the caption of Table~\ref{tab:butterflies_supp_results} for more details. 
}
\centering
\resizebox{1.0\linewidth}{!}{

\begin{tabular}{
>{\columncolor[HTML]{FFFFFF}}c |
>{\columncolor[HTML]{FFFFFF}}l 
>{\columncolor[HTML]{FFFFFF}}l 
>{\columncolor[HTML]{FFFFFF}}l 
>{\columncolor[HTML]{FFFFFF}}l 
>{\columncolor[HTML]{FFFFFF}}l 
>{\columncolor[HTML]{FFFFFF}}l 
>{\columncolor[HTML]{FFFFFF}}l |}
\cline{2-8}
\multicolumn{1}{l|}{\cellcolor[HTML]{FFFFFF}\textbf{}} &
  \multicolumn{7}{c|}{\cellcolor[HTML]{FFFFFF}\textbf{Greebles}} \\ \cline{2-8} 
\multicolumn{1}{l|}{\cellcolor[HTML]{FFFFFF}\textbf{}} &
  \multicolumn{5}{c|}{\cellcolor[HTML]{FFFFFF}Train} &
  \multicolumn{2}{c|}{\cellcolor[HTML]{FFFFFF}Test} \\ \cline{2-8} 
\textbf{} &
  \multicolumn{1}{c|}{\cellcolor[HTML]{FFFFFF}Agara} &
  \multicolumn{1}{c|}{\cellcolor[HTML]{FFFFFF}Bari} &
  \multicolumn{1}{c|}{\cellcolor[HTML]{FFFFFF}Cooka} &
  \multicolumn{1}{c|}{\cellcolor[HTML]{FFFFFF}Micro} &
  \multicolumn{1}{c|}{\cellcolor[HTML]{FFFFFF}Macro} &
  \multicolumn{1}{c|}{\cellcolor[HTML]{FFFFFF}Micro} &
  \multicolumn{1}{c|}{\cellcolor[HTML]{FFFFFF}Macro} \\ \hline
\multicolumn{1}{|c|}{\cellcolor[HTML]{FFFFFF}GT Label$\dagger$} &
  \multicolumn{1}{l|}{\cellcolor[HTML]{FFFFFF}0.51$\pm$0.02} &
  \multicolumn{1}{l|}{\cellcolor[HTML]{FFFFFF}0.37$\pm$0.03} &
  \multicolumn{1}{l|}{\cellcolor[HTML]{FFFFFF}0.43$\pm$0.03} &
  \multicolumn{1}{l|}{\cellcolor[HTML]{FFFFFF}0.45$\pm$0.02} &
  \multicolumn{1}{l|}{\cellcolor[HTML]{FFFFFF}0.44$\pm$0.02} &
  \multicolumn{1}{l|}{\cellcolor[HTML]{FFFFFF}0.50$\pm$0.01} &
  0.49$\pm$0.01 \\ \hline
\multicolumn{1}{|c|}{\cellcolor[HTML]{FFFFFF}$\phi_{static}\dagger$} &
  \multicolumn{1}{l|}{\cellcolor[HTML]{FFFFFF}0.63$\pm$0.03} &
  \multicolumn{1}{l|}{\cellcolor[HTML]{FFFFFF}\textbf{0.43$\pm$0.04}} &
  \multicolumn{1}{l|}{\cellcolor[HTML]{FFFFFF}0.55$\pm$0.05} &
  \multicolumn{1}{l|}{\cellcolor[HTML]{FFFFFF}\textbf{0.55$\pm$0.03}} &
  \multicolumn{1}{l|}{\cellcolor[HTML]{FFFFFF}\textbf{0.53$\pm$0.03}} &
  \multicolumn{1}{l|}{\cellcolor[HTML]{FFFFFF}\textbf{0.64$\pm$0.01}} &
  \textbf{0.61$\pm$0.01} \\ \hline
\multicolumn{1}{|c|}{\cellcolor[HTML]{FFFFFF}$\phi_{static_time}\dagger$} &
  \multicolumn{1}{l|}{\cellcolor[HTML]{FFFFFF}\textbf{0.64$\pm$0.04}} &
  \multicolumn{1}{l|}{\cellcolor[HTML]{FFFFFF}0.39$\pm$0.02} &
  \multicolumn{1}{l|}{\cellcolor[HTML]{FFFFFF}0.54$\pm$0.04} &
  \multicolumn{1}{l|}{\cellcolor[HTML]{FFFFFF}0.54$\pm$0.03} &
  \multicolumn{1}{l|}{\cellcolor[HTML]{FFFFFF}0.52$\pm$0.03} &
  \multicolumn{1}{l|}{\cellcolor[HTML]{FFFFFF}0.61$\pm$0.01} &
  0.59$\pm$0.02 \\ \hline
\multicolumn{1}{|c|}{\cellcolor[HTML]{FFFFFF}$\phi_{dkt}$} &
  \multicolumn{1}{l|}{\cellcolor[HTML]{FFFFFF}0.59$\pm$0.03} &
  \multicolumn{1}{l|}{\cellcolor[HTML]{FFFFFF}0.41$\pm$0.03} &
  \multicolumn{1}{l|}{\cellcolor[HTML]{FFFFFF}0.49$\pm$0.05} &
  \multicolumn{1}{l|}{\cellcolor[HTML]{FFFFFF}0.52$\pm$0.02} &
  \multicolumn{1}{l|}{\cellcolor[HTML]{FFFFFF}0.50$\pm$0.02} &
  \multicolumn{1}{l|}{\cellcolor[HTML]{FFFFFF}0.59$\pm$0.02} &
  0.55$\pm$0.02 \\ \hline
\multicolumn{1}{|c|}{\cellcolor[HTML]{FFFFFF}$\phi_{transformer}$} &
  \multicolumn{1}{l|}{\cellcolor[HTML]{FFFFFF}0.52$\pm$0.11} &
  \multicolumn{1}{l|}{\cellcolor[HTML]{FFFFFF}0.36$\pm$0.03} &
  \multicolumn{1}{l|}{\cellcolor[HTML]{FFFFFF}0.45$\pm$0.05} &
  \multicolumn{1}{l|}{\cellcolor[HTML]{FFFFFF}0.46$\pm$0.07} &
  \multicolumn{1}{l|}{\cellcolor[HTML]{FFFFFF}0.45$\pm$0.06} &
  \multicolumn{1}{l|}{\cellcolor[HTML]{FFFFFF}0.44$\pm$0.08} &
  0.43$\pm$0.08 \\ \hline
\multicolumn{1}{|c|}{\cellcolor[HTML]{FFFFFF}$\phi_{prototype}$} &
  \multicolumn{1}{l|}{\cellcolor[HTML]{FFFFFF}0.58$\pm$0.03} &
  \multicolumn{1}{l|}{\cellcolor[HTML]{FFFFFF}0.42$\pm$0.01} &
  \multicolumn{1}{l|}{\cellcolor[HTML]{FFFFFF}0.54$\pm$0.05} &
  \multicolumn{1}{l|}{\cellcolor[HTML]{FFFFFF}0.52$\pm$0.02} &
  \multicolumn{1}{l|}{\cellcolor[HTML]{FFFFFF}0.51$\pm$0.02} &
  \multicolumn{1}{l|}{\cellcolor[HTML]{FFFFFF}0.58$\pm$0.02} &
  0.57$\pm$0.02 \\ \hline
\multicolumn{1}{|c|}{\cellcolor[HTML]{FFFFFF}$\phi_{exemplar}$} &
  \multicolumn{1}{l|}{\cellcolor[HTML]{FFFFFF}0.59$\pm$0.02} &
  \multicolumn{1}{l|}{\cellcolor[HTML]{FFFFFF}\textbf{0.43$\pm$0.03}} &
  \multicolumn{1}{l|}{\cellcolor[HTML]{FFFFFF}0.52$\pm$0.05} &
  \multicolumn{1}{l|}{\cellcolor[HTML]{FFFFFF}0.52$\pm$0.03} &
  \multicolumn{1}{l|}{\cellcolor[HTML]{FFFFFF}0.51$\pm$0.03} &
  \multicolumn{1}{l|}{\cellcolor[HTML]{FFFFFF}0.63$\pm$0.01} &
  0.60$\pm$0.01 \\ \hline
\multicolumn{1}{|c|}{\cellcolor[HTML]{FFFFFF}$\phi_{direct(base)}$} &
  \multicolumn{1}{l|}{\cellcolor[HTML]{FFFFFF}0.37$\pm$0.02} &
  \multicolumn{1}{l|}{\cellcolor[HTML]{FFFFFF}0.35$\pm$0.02} &
  \multicolumn{1}{l|}{\cellcolor[HTML]{FFFFFF}0.36$\pm$0.01} &
  \multicolumn{1}{l|}{\cellcolor[HTML]{FFFFFF}0.36$\pm$0.00} &
  \multicolumn{1}{l|}{\cellcolor[HTML]{FFFFFF}0.36$\pm$0.00} &
  \multicolumn{1}{l|}{\cellcolor[HTML]{FFFFFF}0.34$\pm$0.02} &
  0.34$\pm$0.02 \\ \hline
\multicolumn{1}{|c|}{\cellcolor[HTML]{FFFFFF}$\phi_{direct(y)}$} &
  \multicolumn{1}{l|}{\cellcolor[HTML]{FFFFFF}0.59$\pm$0.02} &
  \multicolumn{1}{l|}{\cellcolor[HTML]{FFFFFF}0.41$\pm$0.03} &
  \multicolumn{1}{l|}{\cellcolor[HTML]{FFFFFF}0.51$\pm$0.04} &
  \multicolumn{1}{l|}{\cellcolor[HTML]{FFFFFF}0.52$\pm$0.02} &
  \multicolumn{1}{l|}{\cellcolor[HTML]{FFFFFF}0.50$\pm$0.03} &
  \multicolumn{1}{l|}{\cellcolor[HTML]{FFFFFF}0.59$\pm$0.02} &
  0.55$\pm$0.02 \\ \hline
\multicolumn{1}{|c|}{\cellcolor[HTML]{FFFFFF}$\phi_{direct(y, \mathbf{z})}\dagger$} &
  \multicolumn{1}{l|}{\cellcolor[HTML]{FFFFFF}0.62$\pm$0.03} &
  \multicolumn{1}{l|}{\cellcolor[HTML]{FFFFFF}0.42$\pm$0.03} &
  \multicolumn{1}{l|}{\cellcolor[HTML]{FFFFFF}0.55$\pm$0.05} &
  \multicolumn{1}{l|}{\cellcolor[HTML]{FFFFFF}\textbf{0.55$\pm$0.03}} &
  \multicolumn{1}{l|}{\cellcolor[HTML]{FFFFFF}\textbf{0.53$\pm$0.03}} &
  \multicolumn{1}{l|}{\cellcolor[HTML]{FFFFFF}0.60$\pm$0.02} &
  0.57$\pm$0.03 \\ \hline
\multicolumn{1}{|c|}{\cellcolor[HTML]{FFFFFF}$\phi_{cls\_pred(base)}$} &
  \multicolumn{1}{l|}{\cellcolor[HTML]{FFFFFF}0.63$\pm$0.03} &
  \multicolumn{1}{l|}{\cellcolor[HTML]{FFFFFF}0.40$\pm$0.02} &
  \multicolumn{1}{l|}{\cellcolor[HTML]{FFFFFF}\textbf{0.56$\pm$0.06}} &
  \multicolumn{1}{l|}{\cellcolor[HTML]{FFFFFF}\textbf{0.55$\pm$0.02}} &
  \multicolumn{1}{l|}{\cellcolor[HTML]{FFFFFF}\textbf{0.53$\pm$0.03}} &
  \multicolumn{1}{l|}{\cellcolor[HTML]{FFFFFF}0.61$\pm$0.02} &
  0.60$\pm$0.03 \\ \hline
\multicolumn{1}{|c|}{\cellcolor[HTML]{FFFFFF}$\phi_{cls\_pred(y)}\dagger$} &
  \multicolumn{1}{l|}{\cellcolor[HTML]{FFFFFF}0.62$\pm$0.03} &
  \multicolumn{1}{l|}{\cellcolor[HTML]{FFFFFF}0.41$\pm$0.03} &
  \multicolumn{1}{l|}{\cellcolor[HTML]{FFFFFF}0.54$\pm$0.03} &
  \multicolumn{1}{l|}{\cellcolor[HTML]{FFFFFF}0.54$\pm$0.02} &
  \multicolumn{1}{l|}{\cellcolor[HTML]{FFFFFF}0.52$\pm$0.03} &
  \multicolumn{1}{l|}{\cellcolor[HTML]{FFFFFF}0.60$\pm$0.02} &
  0.57$\pm$0.03 \\ \hline
\multicolumn{1}{|c|}{\cellcolor[HTML]{FFFFFF}$\phi_{cls\_pred(y, \mathbf{z})}$} &
  \multicolumn{1}{l|}{\cellcolor[HTML]{FFFFFF}0.63$\pm$0.03} &
  \multicolumn{1}{l|}{\cellcolor[HTML]{FFFFFF}0.41$\pm$0.03} &
  \multicolumn{1}{l|}{\cellcolor[HTML]{FFFFFF}0.55$\pm$0.03} &
  \multicolumn{1}{l|}{\cellcolor[HTML]{FFFFFF}\textbf{0.55$\pm$0.02}} &
  \multicolumn{1}{l|}{\cellcolor[HTML]{FFFFFF}\textbf{0.53$\pm$0.01}} &
  \multicolumn{1}{l|}{\cellcolor[HTML]{FFFFFF}0.61$\pm$0.02} &
  0.59$\pm$0.02 \\ \hline
\end{tabular}
}\label{tab:greebles_supp_results}
\end{table}

\begin{table}[h]
\caption{Performance of models trained using a pre-trained ResNet with partially frozen weights (as described in Sec.~\ref{ResNet_backbone_method}). We only compare model variants that appear in the main text. 
Similar to the original experiment results the classifier prediction model ($\phi_{cls\_pred}$) performs the best. 
However, the overall performance decreases slightly across the board. We observe a larger decrease for the direct response model ($\phi_{direct}$), likely due to the larger dependence that it  has on the feature space.}
\centering
\resizebox{1.0\linewidth}{!}{
\begin{tabular}{
>{\columncolor[HTML]{FFFFFF}}c |
>{\columncolor[HTML]{FFFFFF}}l 
>{\columncolor[HTML]{FFFFFF}}l 
>{\columncolor[HTML]{FFFFFF}}l 
>{\columncolor[HTML]{FFFFFF}}l 
>{\columncolor[HTML]{FFFFFF}}l 
>{\columncolor[HTML]{FFFFFF}}l 
>{\columncolor[HTML]{FFFFFF}}l 
>{\columncolor[HTML]{FFFFFF}}l 
>{\columncolor[HTML]{FFFFFF}}l |}
\cline{2-10}
\multicolumn{1}{l|}{\cellcolor[HTML]{FFFFFF}\textbf{}} &
  \multicolumn{9}{c|}{\cellcolor[HTML]{FFFFFF}\textbf{Butterflies}} \\ \cline{2-10} 
\multicolumn{1}{l|}{\cellcolor[HTML]{FFFFFF}\textbf{}} &
  \multicolumn{7}{c|}{\cellcolor[HTML]{FFFFFF}Train} &
  \multicolumn{2}{c|}{\cellcolor[HTML]{FFFFFF}Test} \\ \cline{2-10} 
\multicolumn{1}{l|}{\cellcolor[HTML]{FFFFFF}\textbf{}} &
  \multicolumn{1}{c|}{\cellcolor[HTML]{FFFFFF}\begin{tabular}[c]{@{}c@{}}Cabbage\\ White\end{tabular}} &
  \multicolumn{1}{l|}{\cellcolor[HTML]{FFFFFF}Monarch} &
  \multicolumn{1}{l|}{\cellcolor[HTML]{FFFFFF}Queen} &
  \multicolumn{1}{c|}{\cellcolor[HTML]{FFFFFF}\begin{tabular}[c]{@{}c@{}}Red\\ Admiral\end{tabular}} &
  \multicolumn{1}{l|}{\cellcolor[HTML]{FFFFFF}Viceroy} &
  \multicolumn{1}{c|}{\cellcolor[HTML]{FFFFFF}Micro} &
  \multicolumn{1}{c|}{\cellcolor[HTML]{FFFFFF}Macro} &
  \multicolumn{1}{c|}{\cellcolor[HTML]{FFFFFF}Micro} &
  \multicolumn{1}{c|}{\cellcolor[HTML]{FFFFFF}Macro} \\ \hline
\multicolumn{1}{|c|}{\cellcolor[HTML]{FFFFFF}GT Label} &
  \multicolumn{1}{l|}{\cellcolor[HTML]{FFFFFF}0.94$\pm$0.03} &
  \multicolumn{1}{l|}{\cellcolor[HTML]{FFFFFF}0.32$\pm$0.01} &
  \multicolumn{1}{l|}{\cellcolor[HTML]{FFFFFF}0.36$\pm$0.04} &
  \multicolumn{1}{l|}{\cellcolor[HTML]{FFFFFF}0.65$\pm$0.04} &
  \multicolumn{1}{l|}{\cellcolor[HTML]{FFFFFF}0.27$\pm$0.01} &
  \multicolumn{1}{l|}{\cellcolor[HTML]{FFFFFF}0.48$\pm$0.02} &
  \multicolumn{1}{l|}{\cellcolor[HTML]{FFFFFF}0.51$\pm$0.02} &
  \multicolumn{1}{l|}{\cellcolor[HTML]{FFFFFF}0.58$\pm$0.02} &
  0.61$\pm$0.02 \\ \hline
\multicolumn{1}{|c|}{\cellcolor[HTML]{FFFFFF}$\phi_{static}$} &
  \multicolumn{1}{l|}{\cellcolor[HTML]{FFFFFF}0.95$\pm$0.01} &
  \multicolumn{1}{l|}{\cellcolor[HTML]{FFFFFF}0.36$\pm$0.02} &
  \multicolumn{1}{l|}{\cellcolor[HTML]{FFFFFF}0.32$\pm$0.04} &
  \multicolumn{1}{l|}{\cellcolor[HTML]{FFFFFF}0.65$\pm$0.04} &
  \multicolumn{1}{l|}{\cellcolor[HTML]{FFFFFF}0.27$\pm$0.03} &
  \multicolumn{1}{l|}{\cellcolor[HTML]{FFFFFF}0.62$\pm$0.02} &
  \multicolumn{1}{l|}{\cellcolor[HTML]{FFFFFF}0.51$\pm$0.02} &
  \multicolumn{1}{l|}{\cellcolor[HTML]{FFFFFF}0.67$\pm$0.03} &
  0.57$\pm$0.03 \\ \hline
\multicolumn{1}{|c|}{\cellcolor[HTML]{FFFFFF}$\phi_{static\_time}$} &
  \multicolumn{1}{l|}{\cellcolor[HTML]{FFFFFF}0.44$\pm$0.02} &
  \multicolumn{1}{l|}{\cellcolor[HTML]{FFFFFF}0.25$\pm$0.02} &
  \multicolumn{1}{l|}{\cellcolor[HTML]{FFFFFF}0.22$\pm$0.01} &
  \multicolumn{1}{l|}{\cellcolor[HTML]{FFFFFF}0.33$\pm$0.03} &
  \multicolumn{1}{l|}{\cellcolor[HTML]{FFFFFF}0.19$\pm$0.02} &
  \multicolumn{1}{l|}{\cellcolor[HTML]{FFFFFF}0.28$\pm$0.01} &
  \multicolumn{1}{l|}{\cellcolor[HTML]{FFFFFF}0.29$\pm$0.00} &
  \multicolumn{1}{l|}{\cellcolor[HTML]{FFFFFF}0.27$\pm$0.02} &
  0.34$\pm$0.02 \\ \hline
\multicolumn{1}{|c|}{\cellcolor[HTML]{FFFFFF}$\phi_{direct}$} &
  \multicolumn{1}{l|}{\cellcolor[HTML]{FFFFFF}\textbf{0.97$\pm$0.01}} &
  \multicolumn{1}{l|}{\cellcolor[HTML]{FFFFFF}\textbf{0.41$\pm$0.03}} &
  \multicolumn{1}{l|}{\cellcolor[HTML]{FFFFFF}0.38$\pm$0.06} &
  \multicolumn{1}{l|}{\cellcolor[HTML]{FFFFFF}0.74$\pm$0.07} &
  \multicolumn{1}{l|}{\cellcolor[HTML]{FFFFFF}0.28$\pm$0.03} &
  \multicolumn{1}{l|}{\cellcolor[HTML]{FFFFFF}0.66$\pm$0.03} &
  \multicolumn{1}{l|}{\cellcolor[HTML]{FFFFFF}0.55$\pm$0.02} &
  \multicolumn{1}{l|}{\cellcolor[HTML]{FFFFFF}0.70$\pm$0.03} &
  0.58$\pm$0.04 \\ \hline
\multicolumn{1}{|c|}{\cellcolor[HTML]{FFFFFF}$\phi_{cls_pred}$} &
  \multicolumn{1}{l|}{\cellcolor[HTML]{FFFFFF}\textbf{0.97$\pm$0.01}} &
  \multicolumn{1}{l|}{\cellcolor[HTML]{FFFFFF}0.39$\pm$0.02} &
  \multicolumn{1}{l|}{\cellcolor[HTML]{FFFFFF}\textbf{0.49$\pm$0.02}} &
  \multicolumn{1}{l|}{\cellcolor[HTML]{FFFFFF}\textbf{0.75$\pm$0.06}} &
  \multicolumn{1}{l|}{\cellcolor[HTML]{FFFFFF}\textbf{0.32$\pm$0.03}} &
  \multicolumn{1}{l|}{\cellcolor[HTML]{FFFFFF}\textbf{0.69$\pm$0.01}} &
  \multicolumn{1}{l|}{\cellcolor[HTML]{FFFFFF}\textbf{0.59$\pm$0.01}} &
  \multicolumn{1}{l|}{\cellcolor[HTML]{FFFFFF}\textbf{0.75$\pm$0.01}} &
  \textbf{0.65$\pm$0.02} \\ \hline
\end{tabular}
}
\label{tab:resnet_experiment}
\end{table}

\begin{table}[h]
\caption{Performance of models after concatenating per-class accuracy information to the input vector for the tracing model~(Sec. \ref{PerClAcc}). We only compare model variants that appear in the main text. We observed a boost for all models.}
\centering
\resizebox{1.0\linewidth}{!}{
\begin{tabular}{c|llll|llll|llll|}
\cline{2-13}
\multicolumn{1}{l|}{} &
  \multicolumn{4}{c|}{\textbf{Greebles}} &
  \multicolumn{4}{c|}{\textbf{Eyes}} &
  \multicolumn{4}{c|}{\textbf{Butterflies}} \\ \cline{2-13} 
\multicolumn{1}{l|}{} &
  \multicolumn{2}{c|}{Train} &
  \multicolumn{2}{c|}{Test} &
  \multicolumn{2}{c|}{Train} &
  \multicolumn{2}{c|}{Test} &
  \multicolumn{2}{c|}{Train} &
  \multicolumn{2}{c|}{Test} \\ \cline{2-13} 
\multicolumn{1}{l|}{} &
  \multicolumn{1}{c|}{Micro} &
  \multicolumn{1}{c|}{Macro} &
  \multicolumn{1}{c|}{Micro} &
  \multicolumn{1}{c|}{Macro} &
  \multicolumn{1}{c|}{Micro} &
  \multicolumn{1}{c|}{Macro} &
  \multicolumn{1}{c|}{Micro} &
  \multicolumn{1}{c|}{Macro} &
  \multicolumn{1}{c|}{Micro} &
  \multicolumn{1}{c|}{Macro} &
  \multicolumn{1}{c|}{Micro} &
  \multicolumn{1}{c|}{Macro} \\ \hline
\multicolumn{1}{|c|}{$\phi_{static}\dagger$} &
  \multicolumn{1}{l|}{0.63} &
  \multicolumn{1}{l|}{0.52} &
  \multicolumn{1}{l|}{0.67} &
  0.58 &
  \multicolumn{1}{l|}{0.60} &
  \multicolumn{1}{l|}{0.59} &
  \multicolumn{1}{l|}{0.67} &
  0.68 &
  \multicolumn{1}{l|}{0.55} &
  \multicolumn{1}{l|}{0.53} &
  \multicolumn{1}{l|}{0.64} &
  0.61 \\ \hline
\multicolumn{1}{|c|}{$\phi_{static+perClAcc}$} &
  \multicolumn{1}{l|}{0.65} &
  \multicolumn{1}{l|}{0.54} &
  \multicolumn{1}{l|}{0.69} &
  0.58 &
  \multicolumn{1}{l|}{0.63} &
  \multicolumn{1}{l|}{0.62} &
  \multicolumn{1}{l|}{0.69} &
  0.70 &
  \multicolumn{1}{l|}{\textbf{0.59}} &
  \multicolumn{1}{l|}{\textbf{0.57}} &
  \multicolumn{1}{l|}{\textbf{0.66}} &
  \textbf{0.65} \\ \hline
\multicolumn{1}{|c|}{$\phi_{direct}\dagger$} &
  \multicolumn{1}{l|}{0.70} &
  \multicolumn{1}{l|}{0.59} &
  \multicolumn{1}{l|}{0.77} &
  0.64 &
  \multicolumn{1}{l|}{0.66} &
  \multicolumn{1}{l|}{0.65} &
  \multicolumn{1}{l|}{0.75} &
  0.74 &
  \multicolumn{1}{l|}{0.55} &
  \multicolumn{1}{l|}{0.53} &
  \multicolumn{1}{l|}{0.60} &
  0.57 \\ \hline
\multicolumn{1}{|c|}{$\phi_{direct+perClAcc}$} &
  \multicolumn{1}{l|}{\textbf{0.71}} &
  \multicolumn{1}{l|}{\textbf{0.62}} &
  \multicolumn{1}{l|}{0.78} &
  \textbf{0.67} &
  \multicolumn{1}{l|}{\textbf{0.69}} &
  \multicolumn{1}{l|}{\textbf{0.69}} &
  \multicolumn{1}{l|}{0.78} &
  0.78 &
  \multicolumn{1}{l|}{0.53} &
  \multicolumn{1}{l|}{0.51} &
  \multicolumn{1}{l|}{0.61} &
  0.57 \\ \hline
\multicolumn{1}{|c|}{$\phi_{cls_pred}\dagger$} &
  \multicolumn{1}{l|}{\textbf{0.71}} &
  \multicolumn{1}{l|}{0.62} &
  \multicolumn{1}{l|}{0.77} &
  0.65 &
  \multicolumn{1}{l|}{0.65} &
  \multicolumn{1}{l|}{0.65} &
  \multicolumn{1}{l|}{0.74} &
  0.74 &
  \multicolumn{1}{l|}{0.54} &
  \multicolumn{1}{l|}{0.52} &
  \multicolumn{1}{l|}{0.60} &
  0.57 \\ \hline
\multicolumn{1}{|c|}{$\phi_{cls\_pred+perClAcc}$} &
  \multicolumn{1}{l|}{\textbf{0.71}} &
  \multicolumn{1}{l|}{0.61} &
  \multicolumn{1}{l|}{\textbf{0.79}} &
  \textbf{0.67} &
  \multicolumn{1}{l|}{\textbf{0.69}} &
  \multicolumn{1}{l|}{\textbf{0.69}} &
  \multicolumn{1}{l|}{\textbf{0.79}} &
  \textbf{0.79} &
  \multicolumn{1}{l|}{0.53} &
  \multicolumn{1}{l|}{0.51} &
  \multicolumn{1}{l|}{0.60} &
  0.58 \\ \hline
\end{tabular}
}
\label{tab:per_class_acc_exp}
\end{table}

\begin{table}[h]
\caption{We train the $\phi_{direct}$ tracing model on the butterflies dataset with different embedding dimensions. We find that embedding dimension has no impact on performance.}
\centering
\resizebox{1.0\linewidth}{!}{
\begin{tabular}{
>{\columncolor[HTML]{FFFFFF}}c |
>{\columncolor[HTML]{FFFFFF}}l 
>{\columncolor[HTML]{FFFFFF}}l 
>{\columncolor[HTML]{FFFFFF}}l 
>{\columncolor[HTML]{FFFFFF}}l 
>{\columncolor[HTML]{FFFFFF}}l 
>{\columncolor[HTML]{FFFFFF}}l 
>{\columncolor[HTML]{FFFFFF}}l 
>{\columncolor[HTML]{FFFFFF}}l 
>{\columncolor[HTML]{FFFFFF}}l |}
\cline{2-10}
\multicolumn{1}{l|}{\cellcolor[HTML]{FFFFFF}\textbf{}} &
  \multicolumn{9}{c|}{\cellcolor[HTML]{FFFFFF}\textbf{Butterflies}} \\ \cline{2-10} 
\multicolumn{1}{l|}{\cellcolor[HTML]{FFFFFF}\textbf{}} &
  \multicolumn{7}{c|}{\cellcolor[HTML]{FFFFFF}Train} &
  \multicolumn{2}{c|}{\cellcolor[HTML]{FFFFFF}Test} \\ \cline{2-10} 
\multicolumn{1}{l|}{\cellcolor[HTML]{FFFFFF}\textbf{}} &
  \multicolumn{1}{c|}{\cellcolor[HTML]{FFFFFF}\begin{tabular}[c]{@{}c@{}}Cabbage\\ White\end{tabular}} &
  \multicolumn{1}{l|}{\cellcolor[HTML]{FFFFFF}Monarch} &
  \multicolumn{1}{l|}{\cellcolor[HTML]{FFFFFF}Queen} &
  \multicolumn{1}{c|}{\cellcolor[HTML]{FFFFFF}\begin{tabular}[c]{@{}c@{}}Red\\ Admiral\end{tabular}} &
  \multicolumn{1}{l|}{\cellcolor[HTML]{FFFFFF}Viceroy} &
  \multicolumn{1}{c|}{\cellcolor[HTML]{FFFFFF}Micro} &
  \multicolumn{1}{c|}{\cellcolor[HTML]{FFFFFF}Macro} &
  \multicolumn{1}{c|}{\cellcolor[HTML]{FFFFFF}Micro} &
  \multicolumn{1}{c|}{\cellcolor[HTML]{FFFFFF}Macro} \\ \hline
\multicolumn{1}{|c|}{\cellcolor[HTML]{FFFFFF}$\phi_{direct\_dim8}$} &
  \multicolumn{1}{l|}{\cellcolor[HTML]{FFFFFF}0.98$\pm$0.01} &
  \multicolumn{1}{l|}{\cellcolor[HTML]{FFFFFF}0.41$\pm$0.03} &
  \multicolumn{1}{l|}{\cellcolor[HTML]{FFFFFF}0.45$\pm$0.06} &
  \multicolumn{1}{l|}{\cellcolor[HTML]{FFFFFF}0.77$\pm$0.07} &
  \multicolumn{1}{l|}{\cellcolor[HTML]{FFFFFF}0.35$\pm$0.03} &
  \multicolumn{1}{l|}{\cellcolor[HTML]{FFFFFF}0.70$\pm$0.03} &
  \multicolumn{1}{l|}{\cellcolor[HTML]{FFFFFF}0.59$\pm$0.03} &
  \multicolumn{1}{l|}{\cellcolor[HTML]{FFFFFF}0.77$\pm$0.03} &
  0.65$\pm$0.04 \\ \hline
\multicolumn{1}{|c|}{\cellcolor[HTML]{FFFFFF}$\phi_{direct\_dim16}$} &
  \multicolumn{1}{l|}{\cellcolor[HTML]{FFFFFF}0.98$\pm$0.01} &
  \multicolumn{1}{l|}{\cellcolor[HTML]{FFFFFF}0.42$\pm$0.05} &
  \multicolumn{1}{l|}{\cellcolor[HTML]{FFFFFF}0.46$\pm$0.04} &
  \multicolumn{1}{l|}{\cellcolor[HTML]{FFFFFF}0.77$\pm$0.07} &
  \multicolumn{1}{l|}{\cellcolor[HTML]{FFFFFF}0.37$\pm$0.03} &
  \multicolumn{1}{l|}{\cellcolor[HTML]{FFFFFF}0.70$\pm$0.02} &
  \multicolumn{1}{l|}{\cellcolor[HTML]{FFFFFF}0.60$\pm$0.02} &
  \multicolumn{1}{l|}{\cellcolor[HTML]{FFFFFF}0.78$\pm$0.03} &
  0.66$\pm$0.03 \\ \hline
\multicolumn{1}{|c|}{\cellcolor[HTML]{FFFFFF}$\phi_{direct\_dim32}$} &
  \multicolumn{1}{l|}{\cellcolor[HTML]{FFFFFF}0.98$\pm$0.01} &
  \multicolumn{1}{l|}{\cellcolor[HTML]{FFFFFF}0.43$\pm$0.04} &
  \multicolumn{1}{l|}{\cellcolor[HTML]{FFFFFF}0.46$\pm$0.03} &
  \multicolumn{1}{l|}{\cellcolor[HTML]{FFFFFF}0.77$\pm$0.06} &
  \multicolumn{1}{l|}{\cellcolor[HTML]{FFFFFF}0.35$\pm$0.02} &
  \multicolumn{1}{l|}{\cellcolor[HTML]{FFFFFF}0.70$\pm$0.02} &
  \multicolumn{1}{l|}{\cellcolor[HTML]{FFFFFF}0.60$\pm$0.01} &
  \multicolumn{1}{l|}{\cellcolor[HTML]{FFFFFF}0.78$\pm$0.02} &
  0.67$\pm$0.02 \\ \hline
\multicolumn{1}{|c|}{\cellcolor[HTML]{FFFFFF}$\phi_{direct\_dim64}$} &
  \multicolumn{1}{l|}{\cellcolor[HTML]{FFFFFF}0.97$\pm$0.01} &
  \multicolumn{1}{l|}{\cellcolor[HTML]{FFFFFF}0.43$\pm$0.03} &
  \multicolumn{1}{l|}{\cellcolor[HTML]{FFFFFF}0.48$\pm$0.04} &
  \multicolumn{1}{l|}{\cellcolor[HTML]{FFFFFF}0.78$\pm$0.06} &
  \multicolumn{1}{l|}{\cellcolor[HTML]{FFFFFF}0.37$\pm$0.03} &
  \multicolumn{1}{l|}{\cellcolor[HTML]{FFFFFF}0.71$\pm$0.02} &
  \multicolumn{1}{l|}{\cellcolor[HTML]{FFFFFF}0.61$\pm$0.02} &
  \multicolumn{1}{l|}{\cellcolor[HTML]{FFFFFF}0.79$\pm$0.01} &
  0.66$\pm$0.01 \\ \hline
\end{tabular}
}
\label{tab:embedding_dimension}
\end{table}
\clearpage
\pagebreak

\renewcommand{\thetable}{B\arabic{table}}
\section{CNN Architecture Details}
\label{cnn_details}
In Table~\ref{tab:backbone} we describe the architecture of the CNN used to encode images for all of the models. 

\begin{table}[h]
\caption{Structure of the CNN backbone used to learn the image representation. The bolded and italicized entries are variable and depend on the experiment and dataset. The number of image channels (\textit{\textbf{img\_chns}}) is three for the Butterflies and Greebles dataset, but is one for Eyes. The Butterflies and OCT datasets contain larger images (144 x 144), and so \textit{\textbf{\text{img\_feats}}} is set to 1296. For the Greebles dataset, the images are (128 x 128) and \textit{\textbf{\text{img\_feats}}} is set to 1204. Finally, the output of the model is the size of the embedding dimension and is set to 16 for all experiments.
}
\centering
\resizebox{.6\linewidth}{!}{
\begin{tabular}{|
>{\columncolor[HTML]{FFFFFF}}c 
>{\columncolor[HTML]{FFFFFF}}c |
>{\columncolor[HTML]{FFFFFF}}c 
>{\columncolor[HTML]{FFFFFF}}c 
>{\columncolor[HTML]{FFFFFF}}c 
>{\columncolor[HTML]{FFFFFF}}c 
>{\columncolor[HTML]{FFFFFF}}c |}
\hline
\multicolumn{2}{|c|}{\cellcolor[HTML]{FFFFFF}\textbf{CNN backbone}} &
  \multicolumn{5}{c|}{\cellcolor[HTML]{FFFFFF}} \\ \hline
\multicolumn{1}{|c|}{\cellcolor[HTML]{FFFFFF}\textbf{layer}} &
  \textbf{in channels} &
  \multicolumn{1}{c|}{\cellcolor[HTML]{FFFFFF}\textbf{out channels}} &
  \multicolumn{1}{c|}{\cellcolor[HTML]{FFFFFF}\textbf{k}} &
  \multicolumn{1}{c|}{\cellcolor[HTML]{FFFFFF}\textbf{s}} &
  \multicolumn{1}{c|}{\cellcolor[HTML]{FFFFFF}\textbf{p}} &
  \textbf{activation} \\ \hline
\multicolumn{1}{|c|}{\cellcolor[HTML]{FFFFFF}conv1} &
  \textit{\textbf{\text{img\_chns}}} &
  \multicolumn{1}{c|}{\cellcolor[HTML]{FFFFFF}8} &
  \multicolumn{1}{c|}{\cellcolor[HTML]{FFFFFF}5} &
  \multicolumn{1}{c|}{\cellcolor[HTML]{FFFFFF}1} &
  \multicolumn{1}{c|}{\cellcolor[HTML]{FFFFFF}2} &
  PReLU \\ \hline
\multicolumn{1}{|c|}{\cellcolor[HTML]{FFFFFF}maxpool1} &
  - &
  \multicolumn{1}{c|}{\cellcolor[HTML]{FFFFFF}-} &
  \multicolumn{1}{c|}{\cellcolor[HTML]{FFFFFF}4} &
  \multicolumn{1}{c|}{\cellcolor[HTML]{FFFFFF}-} &
  \multicolumn{1}{c|}{\cellcolor[HTML]{FFFFFF}-} &
  PReLU \\ \hline
\multicolumn{1}{|c|}{\cellcolor[HTML]{FFFFFF}conv2} &
  8 &
  \multicolumn{1}{c|}{\cellcolor[HTML]{FFFFFF}16} &
  \multicolumn{1}{c|}{\cellcolor[HTML]{FFFFFF}5} &
  \multicolumn{1}{c|}{\cellcolor[HTML]{FFFFFF}1} &
  \multicolumn{1}{c|}{\cellcolor[HTML]{FFFFFF}2} &
  PReLU \\ \hline
\multicolumn{1}{|c|}{\cellcolor[HTML]{FFFFFF}maxpool2} &
  - &
  \multicolumn{1}{c|}{\cellcolor[HTML]{FFFFFF}-} &
  \multicolumn{1}{c|}{\cellcolor[HTML]{FFFFFF}4} &
  \multicolumn{1}{c|}{\cellcolor[HTML]{FFFFFF}-} &
  \multicolumn{1}{c|}{\cellcolor[HTML]{FFFFFF}-} &
  - \\ \hline
\multicolumn{1}{|c|}{\cellcolor[HTML]{FFFFFF}flatten} &
  - &
  \multicolumn{1}{c|}{\cellcolor[HTML]{FFFFFF}-} &
  \multicolumn{1}{c|}{\cellcolor[HTML]{FFFFFF}-} &
  \multicolumn{1}{c|}{\cellcolor[HTML]{FFFFFF}-} &
  \multicolumn{1}{c|}{\cellcolor[HTML]{FFFFFF}-} &
  - \\ \hline
\multicolumn{1}{|c|}{\cellcolor[HTML]{FFFFFF}linear} &
  \textit{\textbf{\text{img\_feats}}} &
  \multicolumn{1}{c|}{\cellcolor[HTML]{FFFFFF}512} &
  \multicolumn{1}{c|}{\cellcolor[HTML]{FFFFFF}-} &
  \multicolumn{1}{c|}{\cellcolor[HTML]{FFFFFF}-} &
  \multicolumn{1}{c|}{\cellcolor[HTML]{FFFFFF}-} &
  PReLU \\ \hline
\multicolumn{1}{|c|}{\cellcolor[HTML]{FFFFFF}linear} &
  512 &
  \multicolumn{1}{c|}{\cellcolor[HTML]{FFFFFF}256} &
  \multicolumn{1}{c|}{\cellcolor[HTML]{FFFFFF}-} &
  \multicolumn{1}{c|}{\cellcolor[HTML]{FFFFFF}-} &
  \multicolumn{1}{c|}{\cellcolor[HTML]{FFFFFF}-} &
  PReLU \\ \hline
\multicolumn{1}{|c|}{\cellcolor[HTML]{FFFFFF}linear} &
  256 &
  \multicolumn{1}{c|}{\cellcolor[HTML]{FFFFFF}256} &
  \multicolumn{1}{c|}{\cellcolor[HTML]{FFFFFF}-} &
  \multicolumn{1}{c|}{\cellcolor[HTML]{FFFFFF}-} &
  \multicolumn{1}{c|}{\cellcolor[HTML]{FFFFFF}-} &
  PReLU \\ \hline
\multicolumn{1}{|c|}{\cellcolor[HTML]{FFFFFF}linear} &
  256 &
  \multicolumn{1}{c|}{\cellcolor[HTML]{FFFFFF} 16} &
  \multicolumn{1}{c|}{\cellcolor[HTML]{FFFFFF}-} &
  \multicolumn{1}{c|}{\cellcolor[HTML]{FFFFFF}-} &
  \multicolumn{1}{c|}{\cellcolor[HTML]{FFFFFF}-} &
  PReLU \\ \hline
\end{tabular}
}
\label{tab:backbone}
\end{table}

\renewcommand{\thefigure}{C\arabic{figure}}
\clearpage
\pagebreak
\section{Additional Results}
\label{additional_results}
We recreate Fig.~4 for all datasets and include results from the Direct Response and Time-Sensitive Model. For the Greebles dataset, we include the histograms of the features of the classes to demonstrate the difficulty of the task.

\begin{figure}[]

    \centering
    \includegraphics[width=1.0\textwidth]{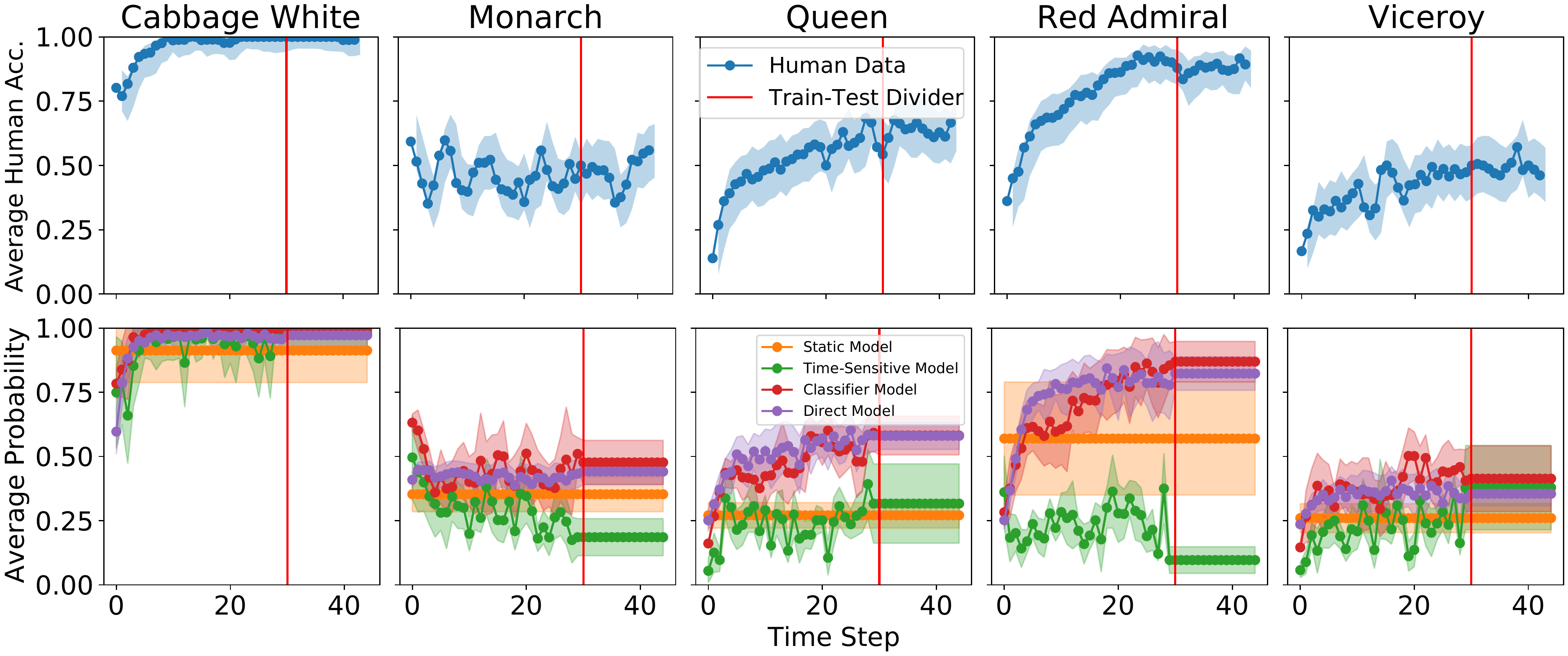}
    \caption{{\bf (Top)} The smoothed average human learner accuracy for each class over time on the Butterflies dataset. The shadowed regions indicate confidence intervals as the number of samples in each time and class bin are not guaranteed to be the same.
    {\bf (Bottom)} The average probability of getting a class correct predicted by the static $\phi_{static}$ model (orange), $\phi_{static\_time}$ model (green), the direct response $\phi_{direct}$ model (purple), and the classifier prediction $\phi_{cls\_pred}$ model (red). At each time-step, for each learner in the test set, the models predict class probabilities for $\sim 50$ images per class. The probabilities are averaged (solid line) and the shadows indicate one standard deviation. While both recurrent models have similar traces, the $\phi_{direct}$ produces smoother average probabilities. 
    }
    \label{fig:supp_ap_butterflies}
\end{figure}

\begin{figure}[t]
    \centering
    \includegraphics[width=1.0\textwidth]{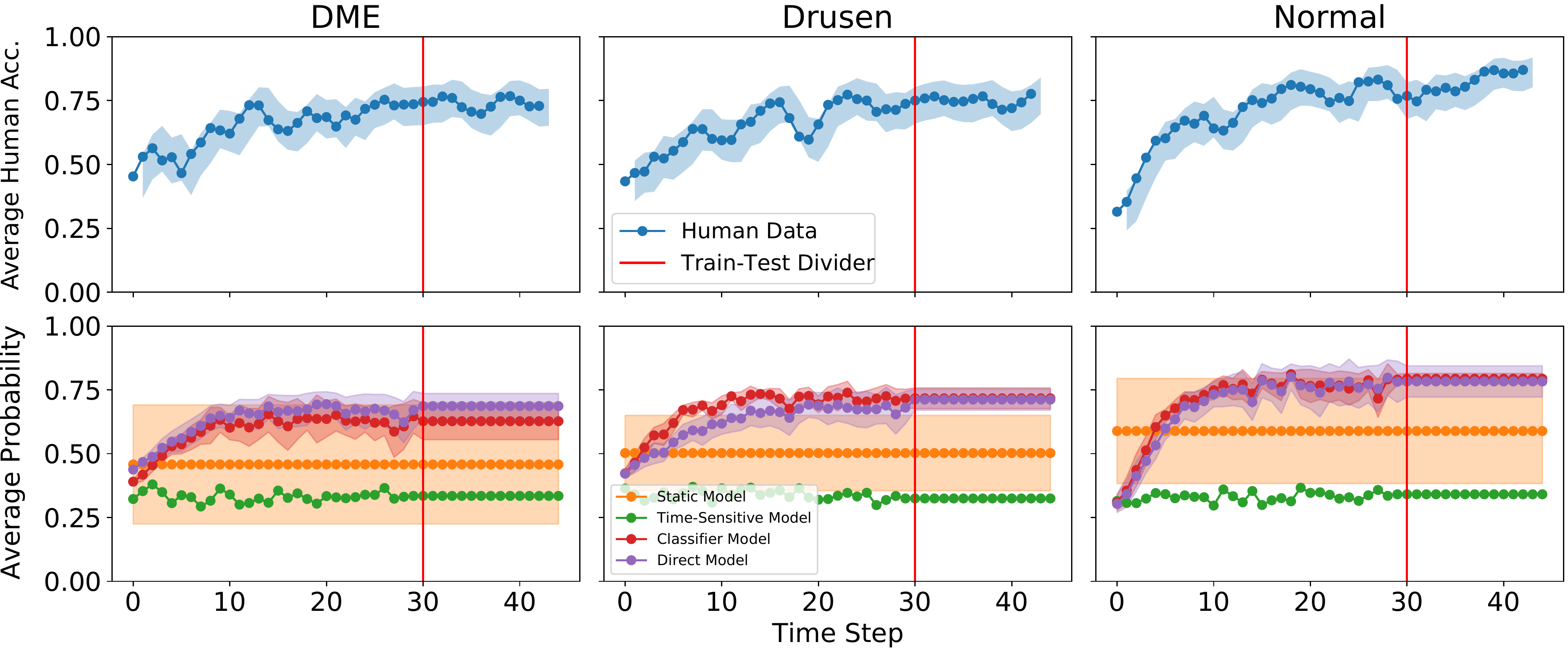}
    \caption{Human and model performance on the Eyes dataset. See Fig.~\ref{fig:supp_ap_butterflies} for a detailed caption.}
    \label{fig:supp_ap_eyes}
\end{figure}

\begin{figure}[t]
    \centering
    \includegraphics[width=0.8\textwidth]{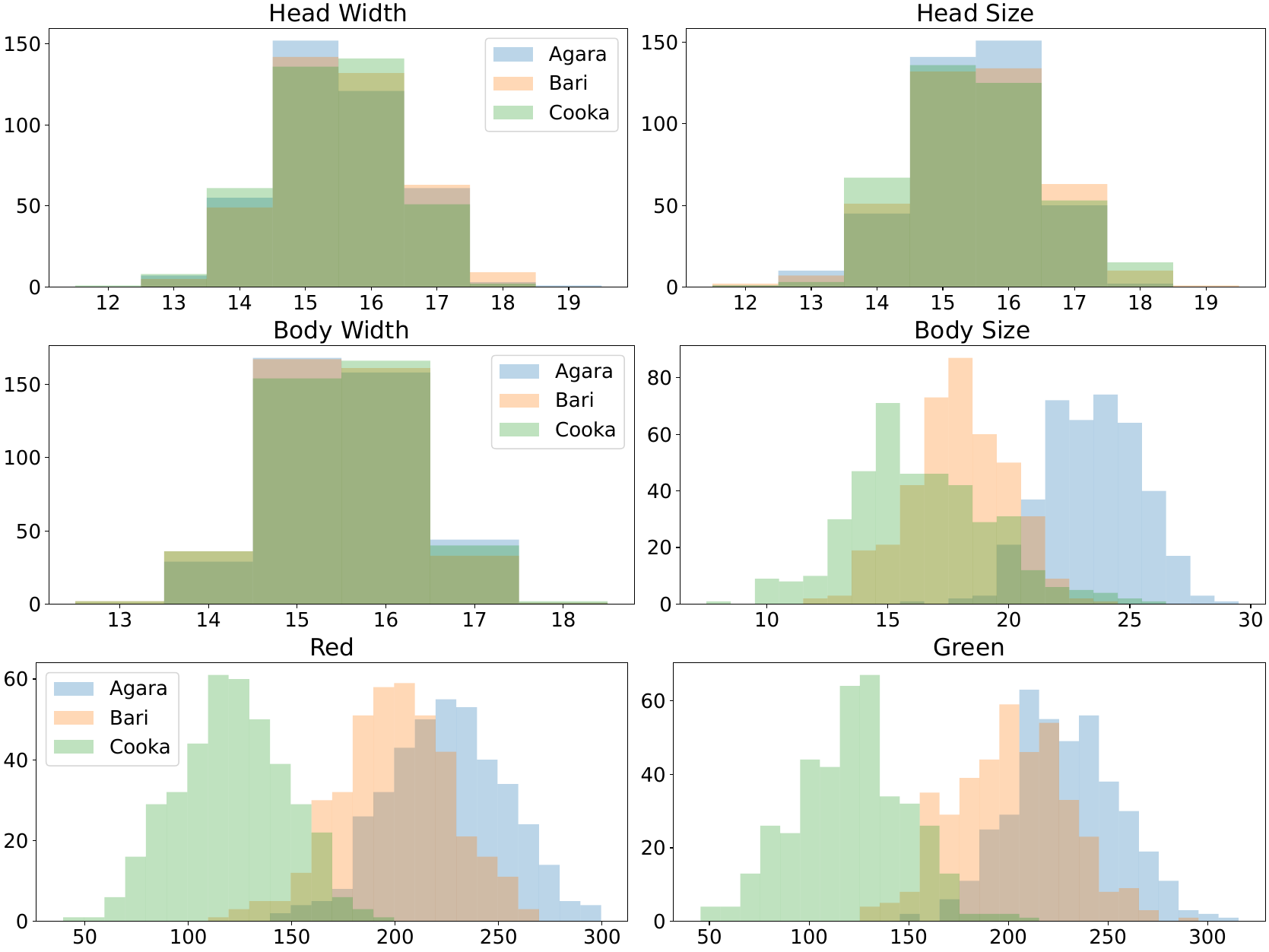}
    \caption{The Greebles dataset was inspired by the one used in~\cite{welinder2010multidimensional}. In our version, the three classes vary in Head Width and Size (top row),  Body Width and Size (middle row), and the Red and Green channel for the RGB color (bottom row). The histograms overlap completely for Head Width, Head Size, and Body Width. These variations serve as distractors since they provide no information about which class an image belongs to. The other features, Body Size, the Red channel, and the Green channel have different distributions and can be used to estimate the class. Agara and Bari are most separable by Body Size, Cooka is most separable from both Agara and Bari in the two color channels. However, note that they are not perfectly separated and it is possible, although less likely, for two images from different classes to take on the same properties. This makes the Greebles dataset particularly challenging, since the important features are both subtle and imperfect for distinguishing between classes. }
    \label{fig:supp_greebles_stats}
\end{figure}

\begin{figure}[t]
    \centering
    \includegraphics[width=1.0\textwidth]{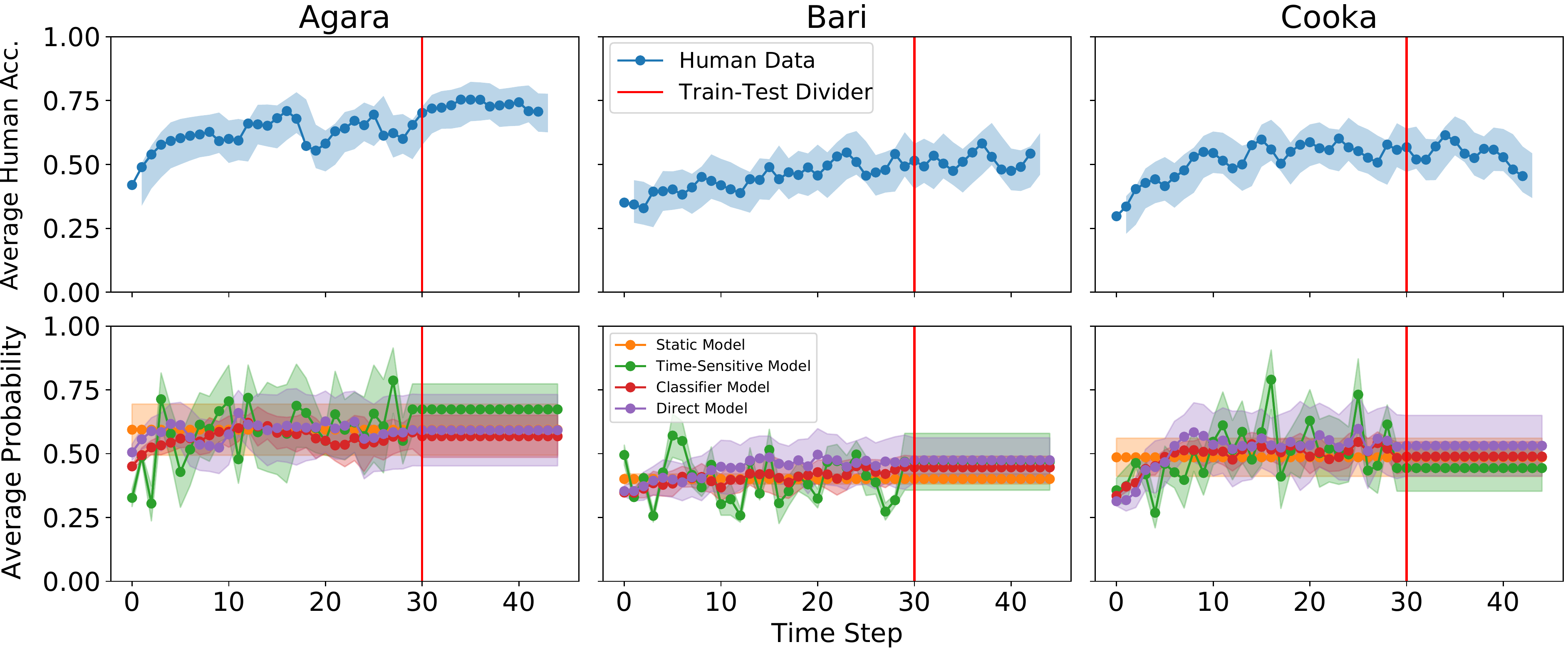}
    \caption{Human and model performance on the Greebles dataset. See Fig.~\ref{fig:supp_ap_butterflies} for a detailed caption.}
    \label{fig:supp_ap_greebles}
\end{figure}

\renewcommand{\thefigure}{D\arabic{figure}}
\clearpage
\pagebreak
\section{Learned Representations}
\label{lstm_representation}
\begin{figure}[]
    \centering
    \includegraphics[width=1.0\textwidth]{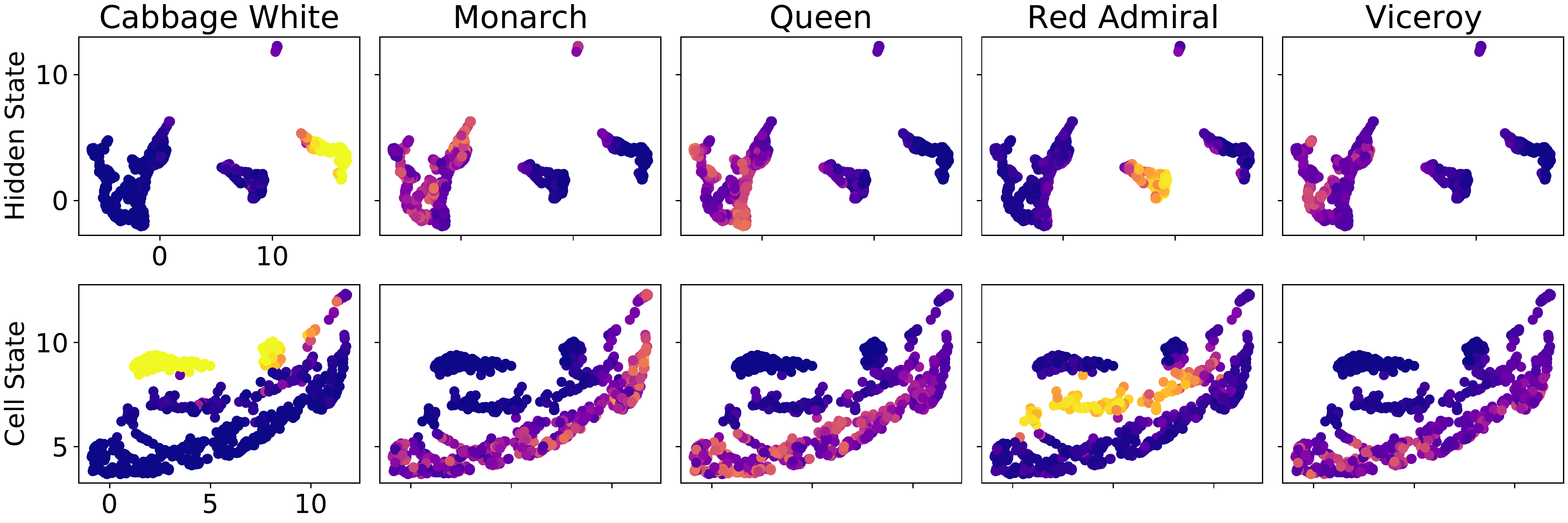}
    \caption{The hidden states and cell states of the LSTM for $\phi_{cls\_pred}$ while tracing 25 test-set learners are plotted in 2D using the UMAP dimensionality reduction algorithm~\cite{mcinnes2018umap-software}.
    {\bf (Top)} The hidden state representations are colored according to the probability (purple to yellow) that a response produced with that hidden state would correctly classify an image of the class in the panel title. We can see that the classes that correspond to the best average performance by humans are in well-defined clusters (\eg Cabbage White), whereas the classes that are commonly mistaken for each other are grouped together and have much weaker probabilities of being correct. {\bf (Bottom)} The cell states are visualized in the same manner. For the cell states, we can see that the clusters seem to be dragged across a single dimension. The Cabbage White and Red Admiral cluster is split in two pieces in the cell state, which we explore in Fig.~\ref{fig:lstm_learner_comparison}.
    }
    \label{fig:lstm_overview}
\end{figure}

\begin{figure}[t]
    \centering
    \includegraphics[width=0.9\textwidth]{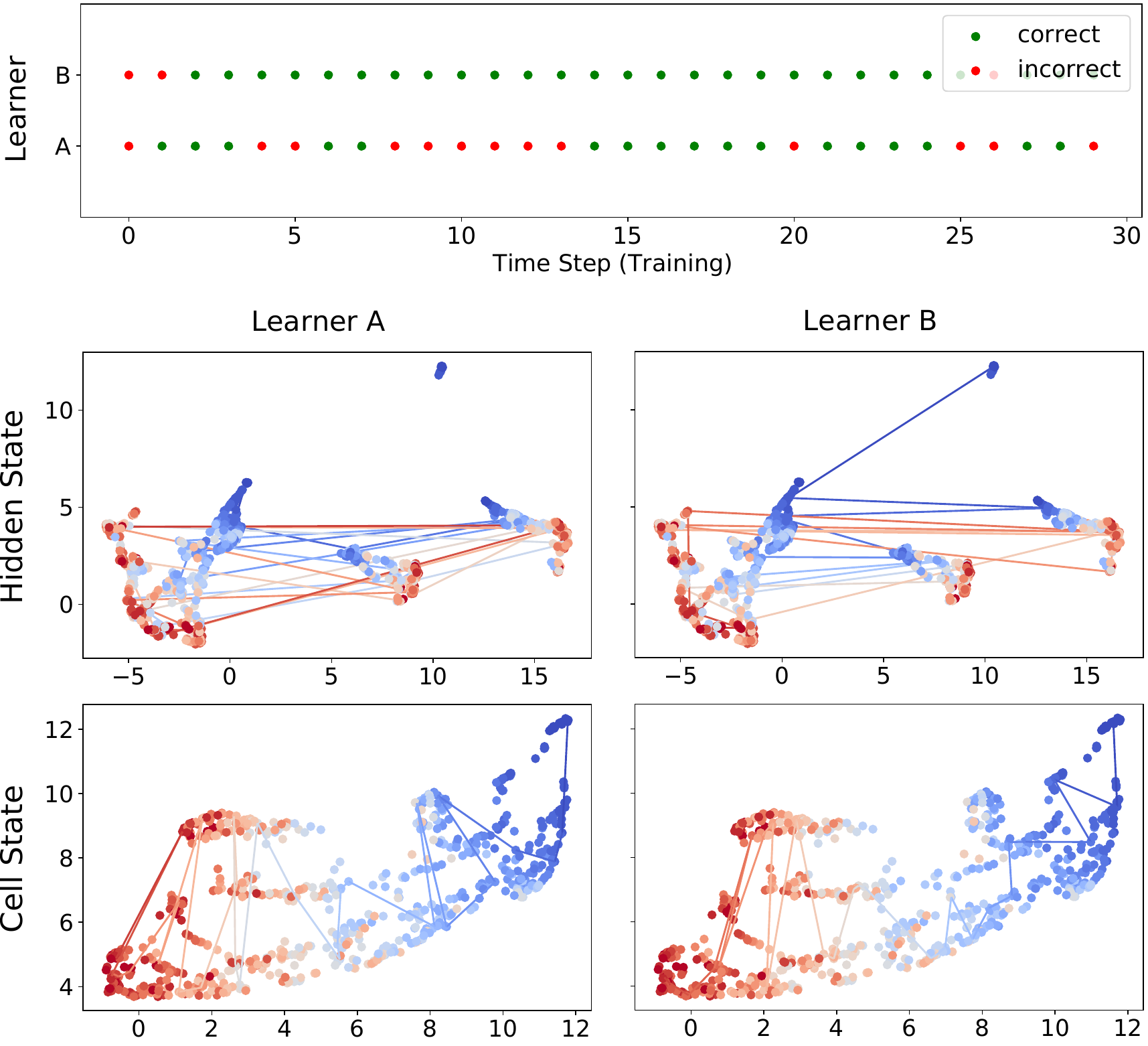}
    \caption{
    {\bf (Top)} The sequence of correct and incorrect responses made by two human learners during training. We selected these two learners as they demonstrate different learning behaviors. It seems Learner B may already be familiar with butterflies. {\bf (Bottom)} We overlay each learners' trajectory through the hidden and cell states. The colors represent the time-step, where dark blue is the beginning of training, light-grey is the middle, and dark red is the end. We see that Learner B's trajectory quickly skips to the left of the cell state, suggesting the LSTM encodes the learners skill level on all classes in certain dimensions of the cell state and uses the hidden state to translate the skill level into an appropriate response for the image shown to the learner.}
    \label{fig:lstm_learner_comparison}
\end{figure}

Here we explore the representations learned by the the classifier prediction model ($\phi_{cls\_pred}$) on the Butterflies dataset. In Fig.~\ref{fig:lstm_overview} we visualize the internal state of the model and in Fig.~\ref{fig:lstm_learner_comparison} we provide an in depth comparison for two different learners. 

\clearpage
\pagebreak
\renewcommand{\thefigure}{E\arabic{figure}}
\section{Feature Space}
\label{feature_space}
\begin{figure}[h]

    \centering
    \includegraphics[width=1.0\textwidth]{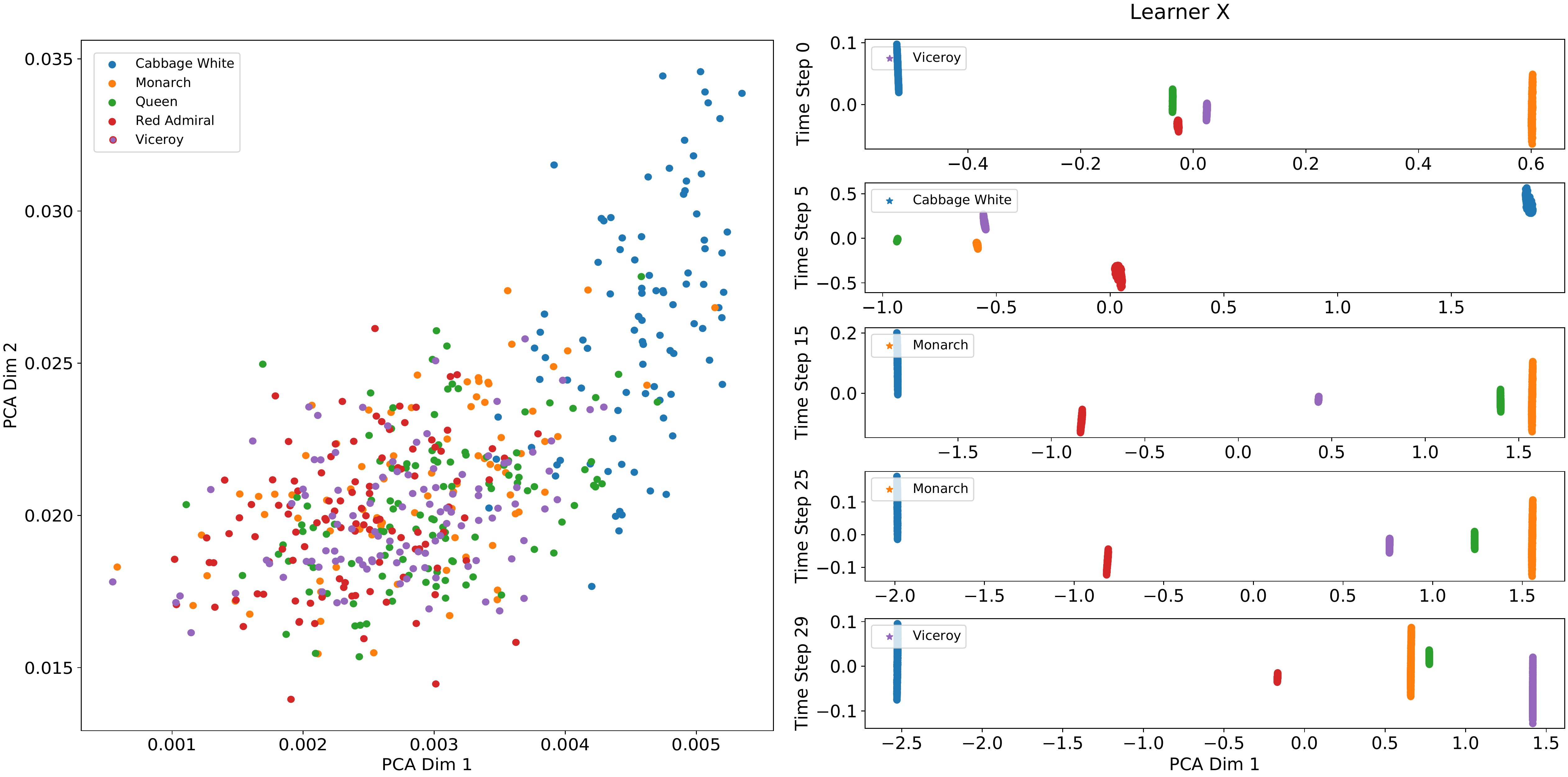}
    \caption{The feature space learned by the CNN must support several types of behaviors, since behavior changes between learners and over time. We use PCA to reduce the learned feature space into two dimensions. \textbf{(Left)} We show a subset of images in the Butterflies dataset colored by the ground truth label. Aside from the Cabbage White class, which is the easiest to identify, the representations are difficult to separate. \textbf{(Right)} We use the hyperplanes predicted while tracing a single learner X to induce a subspace and visualize the features in that subspace. Within the subspaces, the classes are much better separated. Each row shows a subspace induced by a hyperplane for different time-steps - where the time-step is indicated on the left. The colors represent the class and the labelled color is the target class for the image being evaluated in that time-step. We see that, over time, the target class is pushed further to the right and is better separated from the other classes (see orange cluster in time-step 15 vs. 25). Classes that are confused for each other have less separation, whereas classes like Cabbage White, that are rarely confused, are extremely well-separated from the other classes. Also, note that the subspace orientation (target class moved to the right) matches how the dot product between the hyperplanes and features is translated into probabilities in the model.}
    \label{fig:feature_space}
\end{figure}

In Fig.~\ref{fig:feature_space} we visualize the feature space learned by the CNN for the classifier prediction model ($\phi_{cls\_pred}$) on the Butterflies dataset.

\clearpage
\pagebreak

\section{Additional Implementation Details}

\subsection{Types of Learner Responses}
\label{label_ranking}
There are several ways to request information from the learner: they can provide their best guess, a ranked list of guesses, or confidence scores for each class. In these datasets, we ask for a ranking of each learner's top 3 classes as a balance between time-spent and informativeness. While our models are trained on their top choice (equivalent to their best guess), we hypothesize that the extra information available in the ranked responses can be leveraged to improve response prediction performance. We leave this to future work. 

\subsection{Recurrent Neural Networks}
\label{RNN_Details}
Here we elaborate on the details of the recurrent neural network based models. 

\textbf{Direct Response Model.}
At each time-step, this model receives the hidden states, cell states, the learner's response to the previous interaction, the embedding of the current image, and the true label of the current image. The model predicts the response of the learner with respect to the input image. 

\textbf{Classifier Prediction Model.}
At each time-step, this model receives the hidden state, cell state, the embedding of the image from the previous interaction, the learner's response to the previous interaction, and the true label of the current image. The model predicts a classifier that is used to classify the embedding of the input image such that it matches the response of the learner to that input image at that time-step.

Ground truth labels and learner responses are represented as one-hot-encoded vectors. For both models, at the first time-step, some of the values that make up the input vector are not available to the model. For example, the hidden state, cell state, the response to previous interaction, etc. These vectors are initialized to zeros.

\end{document}